\documentclass[%
 reprint,
 amsmath,amssymb,
 aps,
]{revtex4-2}

\usepackage{graphicx}
\usepackage{dcolumn}
\usepackage{bm}

\newcommand\qavg[1]{\overline{q_{#1}}}
\newcommand\qavgavg{\overline{q}}
\newcommand\routside{\rho_O}
\newcommand\rinside{\rho_I}
\newcommand\rhvac{\rho_H}
\newcommand\otherheat{\dot{Q}^O}

\graphicspath{{Images/}}
\begin{document}


\title{Mobility Map Inference from Thermal Modeling of a Building}

\author{Risul Islam}
 \altaffiliation[Also at ]{CS Department, University of California, Riverside, CA, 92521}
 \email{risla002@ucr.edu}

\author{Andrey Lokhov}%
 \email{lokhov@lanl.gov}
\affiliation{%
 Los Alamos National Laboratory, Los Alamos, NM, 87545
}%

\author{Nathan Lemons}%
 \email{nlemons@lanl.gov}
\affiliation{%
 Los Alamos National Laboratory, Los Alamos, NM, 87545
}%

 \author{Michalis Faloutsos}
 \affiliation{%
 University of California, Riverside, CA, 92521
}
\email{michalis@cs.ucr.edu}

\date{\today}
\begin{abstract}
We consider the problem of inferring the mobility map, which is the distribution of the building occupants at each time stamp,  from the temperatures of the rooms. We also want to explore the effects of noise in the temperature measurement, room layout etc. in the reconstruction of the movement of people within the building. Our proposed algorithm tackles down the aforementioned challenges leveraging a parameter learner, the modified Least Square Estimator. In the absence of a complete data set with mobility map, room and ambient temperatures and HVAC data in the public domain, we simulate a physics-based thermal model of the rooms in a building and evaluate the performance of our inference algorithm on this simulated data. We find an upper bound of the noise standard deviation ($\eta_{std} \leq 1F$) in the input temperature data of our model. Within this bound, our algorithm can reconstruct the mobility map with reasonable reconstruction error. Our work can be used in wide range of applications, for example, ensuring the physical security of office buildings, elderly and infant monitoring, building resources management, emergency building evacuation, and vulnerability assessment of HVAC data. Our work brings together multiple research areas, Thermal Modeling and Parameter Estimation, towards achieving a common goal of inferring the distribution of people within a large office building.

\textbf{\textit{Index terms- }Physics model, Thermal modeling, Mobility map, Inference, Least square estimator, Parameter learning, Building identification.}

\end{abstract}

\maketitle
\section{Introduction}\label{sec:1}

\textbf{Research Motivation: }Inferring the mobility map of the building occupants from sensor data has recently become a topic of interest. Modern buildings typically have sophisticated HVAC (Heating, Ventilation, and Air Conditioning) systems which can collect temperature data at relatively fine spatial and temporal scales, making such data a natural place to start when trying to infer the locations of building occupants.  The question emerges:  ``Is the temperature data enough to locate the occupants as they move around the  building?" Such questions can broadly be classed with a wide range of inference questions such as ``By hearing the sound made by a 3D printer, can one reconstruct the object being printed?'' \cite{faruque2016acoustic} and ``Can
we infer the shape of the drum head just from its sound?'' \cite{kac1966can}.  
Unsurprisingly, the answer to our question can be succinctly described as ``it depends'' on the amount of noise in the system.  We build simple, physical models for the temperature of rooms throughout a large building as building occupants move around the building.  These models are used to simulate temperature (and related HVAC) data, which is then fed into an inference algorithm to determine the locations of the building occupants over time.  We also explore the effect of various realistic model parameters such as noise in temperature measurement under which reliable reconstruction is possible. 
Certainly, the application of this work is multi dimensional. For instance, knowledge about real-time location of the occupants can help in the management of resources like printer, oxygen supply, sitting arrangements etc. in smart buildings. Occupancy learning can be used to control the HVAC equipment efficiently to reduce the energy consumption while
maintaining thermal comfort. The most important use is in evacuating during an emergency scenario like an earthquake or radioactive poisoning etc. \cite{zhou2014scanme}. Moreover, knowing the location of occupants inside a building can immediately assist in childcare and elderly health monitoring \cite{horng2011enhancing}. In addition,  inferring the mobility map can help in building’s cyber-physical vulnerability assessment, for instance, understanding whether the HVAC system is compromised. 

{\bf Research Questions. } The research question that we answer in this work are as follows:

(i) Is it possible to infer the mobility map only from the temperatures in an unsupervised fashion?

(ii) How well does the inference algorithm function in the presence of noisy temperature measurement?

\textbf{Related Works:} Despite having potential applications, relatively little research has been done in the field of statistical inference with building HVAC data. The research approaches that have been followed to count the occupants can be divided into the following two broad categories: (i) Classical prediction based approaches, and (ii) System identification based approaches.

(i) {\it Classical prediction based approaches:} This class of approaches mainly utilizes the machine learning algorithms. Most of these works incorporate the data-driven models. Classical prediction based approaches can be categorized further into three types: (i) supervised, (ii) unsupervised, and (iii) combination of supervised and unsupervised approaches.

Supervised approaches make use of the traditional prediction models such as Decision tree, SVM, Multi-layer perceptron, DNN, KNN, Random forest etc.  Several recent works focus on binary occupancy detection (occupied or unoccupied) from Advanced Metering Infrastructure (AMI) data based on deep neural network architectures- CNN and LSTM \cite{feng2020deep} and Genetic algorithm \cite{razavi2019occupancy}. Though these works are very challenging, efficient building management requires the full occupancy map. None of the above mentioned works consider the inference of mobility map inside the building. Ref.  \cite{armknecht2019privacy} tries the Naive Bayes Classifier to predict if there is any occupancy in the room
as well as the occupancy count up to three. Other efforts \cite{golestan2018data, ardakanian2016non, arief2017sd, ghai2012occupancy, fiebig2017detecting} focus on counting any number of occupants in the rooms. These works utilize the data from different sources such as $CO_2$, VOC (Volatile Organic Compounds), room temperature, air flow, humidity, reheat etc. sensors and count up to any number of occupants in the room leveraging the Particle Filters and time series neural network \cite{golestan2018data}, general time series analysis algorithms \cite{ardakanian2016non} , Seasonal Decomposition \cite{arief2017sd}, J48 \cite{ghai2012occupancy}, KNN, RF, MLP, LDA \cite{fiebig2017detecting, szczurek2017occupancy}, and regression model \cite{raykov2016predicting}. Some works \cite{yang2014systematic, candanedo2016accurate} suggested that only temperature data is not sufficient for predicting accurate occupancy pattern but they drive their work in supervised approach. The supervised models count the occupant numbers with a very good accuracy but fails in the absence of pre-acquired training data which is a common flaw for all supervised data-driven models. Semi-supervised and a combination of supervised and unsupervised models are also proposed \cite{arief2017hoc, peng2018using} but they still depend on the training dataset.

To alleviate the problems of supervised models, many recent efforts use unsupervised models to identify occupancy distribution \cite{song2019time, sebastinwolf2019markov, depatla2015occupancy, jain2016software}. The authors of \cite{depatla2015occupancy} are interested in finding the occupancy distribution using the Kullback-Leibler divergence on WiFi received signal strength indicator (RSSI) measurements between a pair of stationary transmitter/receiver antennas.  
Another work \cite{song2019time} utilizes the Hidden Markov Model (HMM) to handle the binary occupancy detection problem (occupied or unoccupied). Utilizing the variations of Hidden Markov Models and other probabilistic models is actually quite common in this research domain \cite{sebastinwolf2019markov, candanedo2017methodology, pedersen2017method, chen2017environmental, ryu2016development, chaney2016evidence}. But these works focus on either binary occupancy detection \cite{candanedo2017methodology, pedersen2017method, chaney2016evidence} or the estimation of the occupancy level (i.e. zero, low, medium and high) \cite{chen2017environmental, chen2017building} or the prediction of activity status of the occupants rather than the actual occupancy count \cite{sebastinwolf2019markov}. 

None of the aforementioned supervised and unsupervised works discuss the effect of noise from data sensing on the models since all of the above-mentioned works follow data-driven architecture which depend on data fusion from various sources, for example, PIR, $CO_2$ concentration, air flow, RFID, temperature sensors, WiFi-signal transmitter and receiver, cameras, ambient vibration, mobile phone etc \cite{ ,pan2014boes, barbour2019planning, szczurek2017occupancy}. Data fusion from these sources can also be challenging \cite{ardakanian2016non}.  These problems can be limited by incorporating a physical model that captures the basic relationships between the occupancy and indoor climate which is our focus in this paper.

{\it (ii) System identification based approaches:} System identification techniques are concerned with using statistical methods to construct mathematical models of dynamical systems. Most of the previous works of this category focus on modeling room temperatures in the presence of HVAC utilizing various strategies, for example, Monte-Carlo simulation \cite{jacob2010optimizing, hu2016building}, Kalman filters \cite{burger2016recursive}, Evolutionary algorithms \cite{ji2016estimating}, and Least Square Estimator \cite{dewson1993least}. While these works concentrate on solely building parameter identification, none of them actually consider finding the occupancy distribution. However, several works focus on both building identification and  occupancy counting \cite{sangogboye2017performance, gruber2014co2, ebadat2013estimation}. Authors of \cite{sangogboye2017performance} incorporated the identification of building parameters using both data-driven and physical models as well as occupancy count based on the data from $CO_2$ concentration and other sensors. But their main focus was performance comparison of occupancy estimation methods based on dedicated vs. common sensors. Ref. \cite{gruber2014co2} investigates a simplified control system by which occupancy is estimated using $CO_2$ concentration sensor responses. The authors also analyzes  the effect of, first, uncertainties in the accordance between $CO_2$ concentration sensor responses and the number of people (i.e. the aspect of estimation errors), and second, latency of the $CO_2$ concentration sensor responses (i.e. the aspect of time delay). The work of \cite{ebadat2013estimation} addresses the problem of estimating the occupancy levels using $CO_2$ concentration again. 

Though these works seem related to our work, there are some fundamental differences: First, all of these works consider the $CO_2$ concentration sensor data, which is a direct indication of the presence of people in a room, as as a function of number of occupants. But our work is more sophisticated and challenging because we consider the temperature as a function of number of people. Actually, the temperature is a convoluted contribution from different sources such as HVAC, external environment, internal devices and, of course, the occupants. Second, none of these works consider the effect of noises in sensor data measurement because the measurements from the sensors usually contain noises (Gaussian) \cite{jung2019human, chen2018building}. We consider the presence of noise in the system in our work and analyze its effect on the performance of our algorithm.

We use the Least Square Estimator (LSE) in the process of identification of our noisy linear system.  LSE has been a well established method in the identification domain. For example, authors in \cite{lokhov2018online} use an extended convex Least Square Estimators to regenerate the dynamic state matrix of transmission power grids from PMU (Phasor Measurement Unit) measurements with timestamp in the presence of ambient fluctuations. Ref \cite{simchowitz2018learning} analyzes the performance of the Ordinary Least Square Estimator for the estimation of linear dynamics $X_{t+1} = AX_t + \eta_t$ from a single trajectory $X_0, X_1, . . . , X_T$ showing that more unstable linear systems are easier to estimate. Another work of \cite{sarkar2019near}, using Least Squares method,  deduces the finite time error bounds for estimating general linear time-invariant (LTI) systems with noise. Motivated by the success of the above mentioned works, our mob Reconstruction Algorithm, based on modified Least Square Estimator with Regularization, is best suited for our problem domain assuming that the sensors' data measuring the temperature of the rooms and environment contain mainly Gaussian noise.

{\bf Contributions. }The main contributions of our work are as follows:

(i) We propose a simple but highly effective unsupervised algorithm to infer the mobility map from only the temperatures considering the system as a noisy linear system.

(ii) We analyze the conditions and constraints of different parameters, for example, room layout, noise in temperature measurement etc. under which mobility map reconstruction is possible.

(iii) We evaluate the performance of our algorithm utilizing a physics-based thermal model which we build in the absence of ground truth.

Note that we do not seek the best algorithm for occupancy estimation; rather we answer whether it is possible to infer mobility map from the temperature considering the noises from temperature measurements.
The rest of this paper is organized as follows: in Section \ref{sec:2} we formulate the thermal model within the various rooms of a large office building and define the mobility map reconstruction problem; in Section \ref{sec:3} we describe our inference method; in Section \ref{sec:4} we provide an empirical assessment of the performance of our algorithm on simulated data and illustrate our approach on two test data sets. In Section \ref{sec:5} we conclude by discussing possible extensions of our method and state some open problems.


\section{\label{sec:2}Thermal Model}
The temperature, $T_i$, in room $i$ can be modeled as a simple linear equation
\begin{equation}\label{eq:1}
\frac{\partial T_i}{\partial t} = \sum_j \rho_{ij}(T_j(t)-T_i(t)) + \sum_k \dot{Q}_{ik}(t)\;,
\end{equation}

\begin{figure}[htp]
    \centering
    \includegraphics[width=9cm]{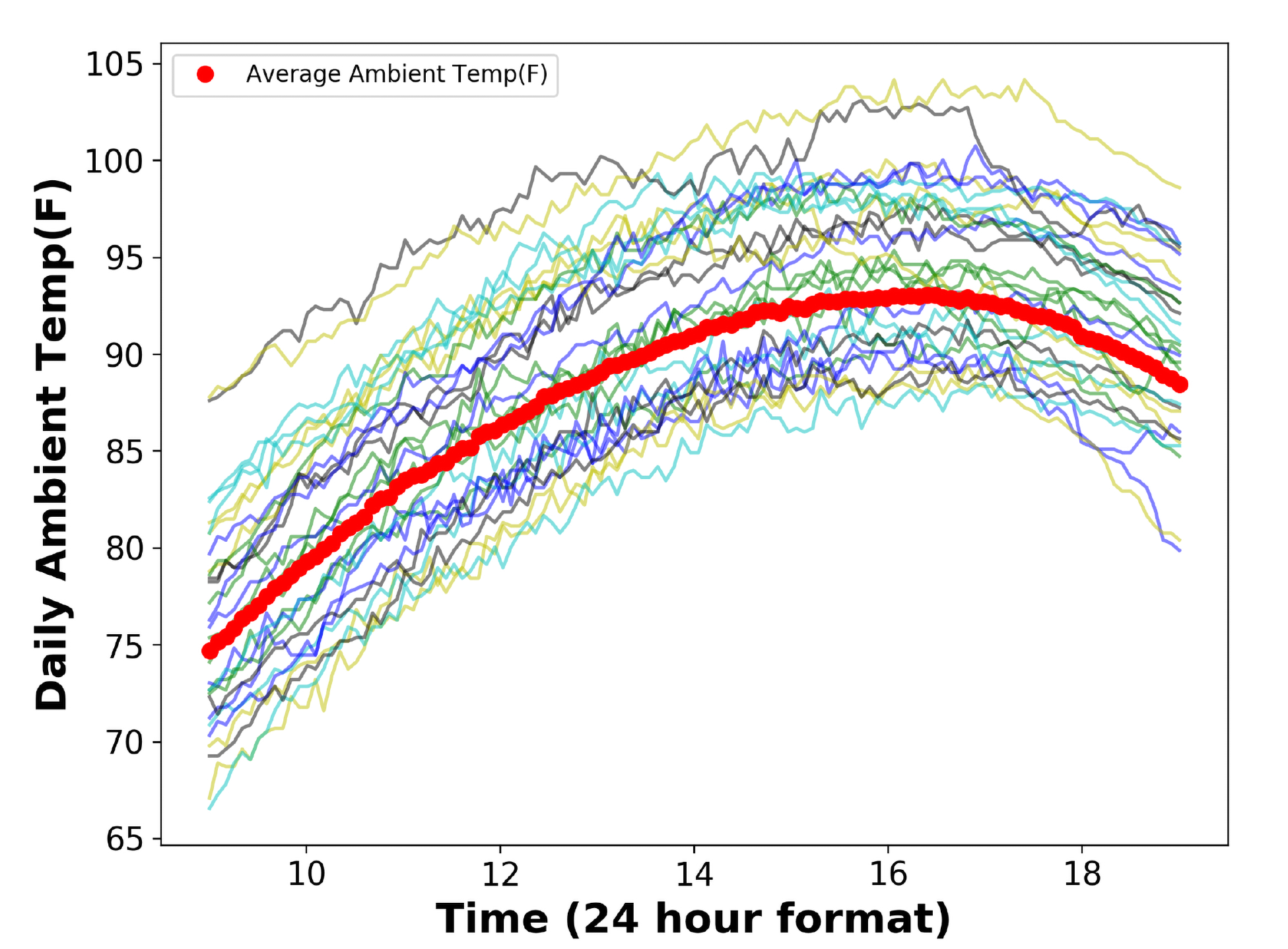}
    \caption{July,2019 daily ambient external temperature data in Marceed, CA. Average daily temperature in Marceed, CA in red dots. }
    \label{fig:tex}
\end{figure}

governed by the temperatures and relative conductances, $T_j$ and $\rho_{ij}$, of each neighboring space $j$, together with the contributions of heat sources $\dot{Q}_{ik}$ such as monitors, computers, and people in room $i$.  
In this setting, the environment outside the building can be considered a neighboring space to those rooms on the building perimeter.  
Likewise, the building heating and cooling system is treated as a neighboring space for each room: its contribution to room $i$ is 
$$\rho_{H}(T_{i}^H(t)-T_i(t))u_i(t)\;,$$
where the additional term $u_i(t)$ is a binary variable representing the status of the system in room $i$ (i.e. ``on'' or ``off'').  
We assume the amount of air piped into each room is fixed if the system is on (and otherwise zero).  Note that mathematically, this is equivalent to assuming that the temperature of the air from the HVAC is fixed and that some rooms receive more air and others less.  (This second formulation better reflects how  HVAC systems actually work in large office buildings.)  Finally, to simplify, we write $\sum_k \dot{Q}_{ik}(t)$ as the sum of two terms, $\dot{Q_{i}^M}$ and $\dot{Q}^O_i$, the heat from people in the room and the heat from all other sources including computers, space heaters, and monitors.
For simplicity, we assume the quantities $\rho_{ij}$ are all the same (and denoted $\rho_I$) when $i$ and $j$ are two rooms of the building.  Otherwise, if one of the indices references the outside environment, we again assume that such $\rho_{ij}$ are all equal to a fixed $\rho_O$.  (Thus each $\rho_{ij}$ is one of three possibilities: $\rho_I$, $\rho_{O}$ or $\rho_H$.)

As we mentioned above, the goal of this paper is to infer the mobility of people in the building from temperature and HVAC data.  However, even our simple model implies other information would be necessary: the building layout, external temperature, and the heat sources $\dot{Q}_{ik}$. Unfortunately, we do not know of any labeled data set that containing (some of) these together with the ground truth movements of the building occupants.  
To compensate for the lack of labeled data, we generate a synthetic data set on which to test our inference methods.  This data is generated as follows.

\textbf{(i) Generation of outside temperature $T^{O}(t)$}: As we suggested above, we assume that the external temperature is constant outside the building, that is we do not take into account possible variances among the building faces such as those due to relative amounts of sun exposure.  In addition, we have focused on cooling as opposed to heating; we simulate an environment where the ambient temperature is above the standard comfort zone of an office building.   In particular we used historical temperature data for Marceed, California during summer time.  We use the average temperature at $5$ minute intervals (taken over a month of data) to simulate data for one 10 hour work day.  Fig. \ref{fig:tex} depicts the whole July, 2019 data from 9:00 to 19:00 as well as our average ambient temperature at each time step. The mean and standard deviation of the average data are 85F and 4.33 respectively. The data is publicly available~\cite{bell2013us}.

\textbf{(ii) Generation of $\dot{Q}^{O}$}: Recall that this term models the heat generation (excluding people) from inside the rooms.  This term is assumed to be constant across each room.  Typically, we can see 3 computers, 1 printer, 4 light bulbs in a room. The temperature contribution by these devices is presented in Section \ref{subsec:supli1}.  Other heating sources, such as Heaters, may also be present. However, as we are simulating an environment to be cooled, we can safely ignore this possibility in the present simulation.  Nonetheless, our map Reconstruction Algorithm, discussed in Section \ref{sec:3}, can also be used to detect other possible heating sources if present.

\textbf{(iii) Generation of $\dot{Q}_i^{M}(t)$}: This term refers to the heat generation in room $i$ due to the presence of the occupants.  We assume that the number of people in a room is \textit{constant} for at least an hour and that (only) on the hour people move stochastically across the building. In particular, we assume that after people come inside an office room, they do not go out at least for an hour. We call this minimum amount of time a person can stay in a room the \textit{Time Range\textbf{(TR)}}. Our modeling choices for the Time Range are further discussed in \ref{sec:4}. Finally, we assume that total number of people in the building is fixed over the entire day.  We denote the total number of people in the building by $N$. Thus, if there are $k$ rooms in the building and the number of people in room $i$ at time $t$ is $n_i(t)$ then:
\[
    \forall t,\;\sum_{i=1}^{k} n_i (t) = N\\
\]


Now, the heat contribution from the occupants in room $i$ is given by, 
$$\dot{Q}_i^{M}(t) = \sum_{j=1}^{n_i(t)}\qavg{j}$$
where, $\qavg{j}$  models the thermal energy emitted on average by person $j$.  In particular, we sample independently the random variables $\qavg{j}$ from a Guassian with mean $\mu = 110W$ and standard deviation $\sigma=1$.

\textbf{(iv) Generation of $n_i(t)$}: We differentiate between large rooms (such as big conference rooms), medium sized rooms (small conference rooms) and small rooms (e.g. single offices).  
By design, more people can be simultaneously present in a large room than in a small room.  To capture this variation we give each room $i$ a weight $w_i$ and stochastically assign each of the $N$ people independently to room $i$ with probability
$$p_i :=\frac{w_i}{\sum_j w_j}$$


Each weight $w_i$ is one of three values (modeling small, medium, or large rooms); the choice of these values can be found in \ref{subsec:supli2}.

\textbf{(v) Generation of $T^{H}_i$}: We have assumed that the HVAC supplies cool air with temperature \textit{constant} over the times but variable across rooms. For instance, in our simulation, we have considered $T_{H} = 50F$ for the big (conference) rooms and $55F$ for smaller conference rooms and offices.

\textbf{(vi) Generation of $u_i(t)$}: This term refers to the ON/OFF state of HVAC's AC system. Depending on the current temperature of a room, the HVAC can trigger the AC system to turn $u_i$ from off to on. We model a simple ``bang-bang'' control which is governed by the setpoints $T_{max}$ and $T_{min}$ which are set to the minimum and maximum temperatures comfortably tolerated by the building occupants. We assume these setpoints are constant across all the rooms in the building.  Given these setpoints, $u_i(t)$ is defined by
\[
    u_i(t)= 
\begin{cases}
    1,& \text{if } T_i\geq T_{max}\;,\\
    0,& \text{if } T_i\leq T_{min}\;,\\
    u_i(t-1),              & \text{otherwise}\;.
\end{cases}
\]

Note that each room will have temperatures fluctuating (mostly) within the range $[T_{min} , T_{max}]$.

\begin{figure}[t]
    \centering
    \includegraphics[width=9cm]{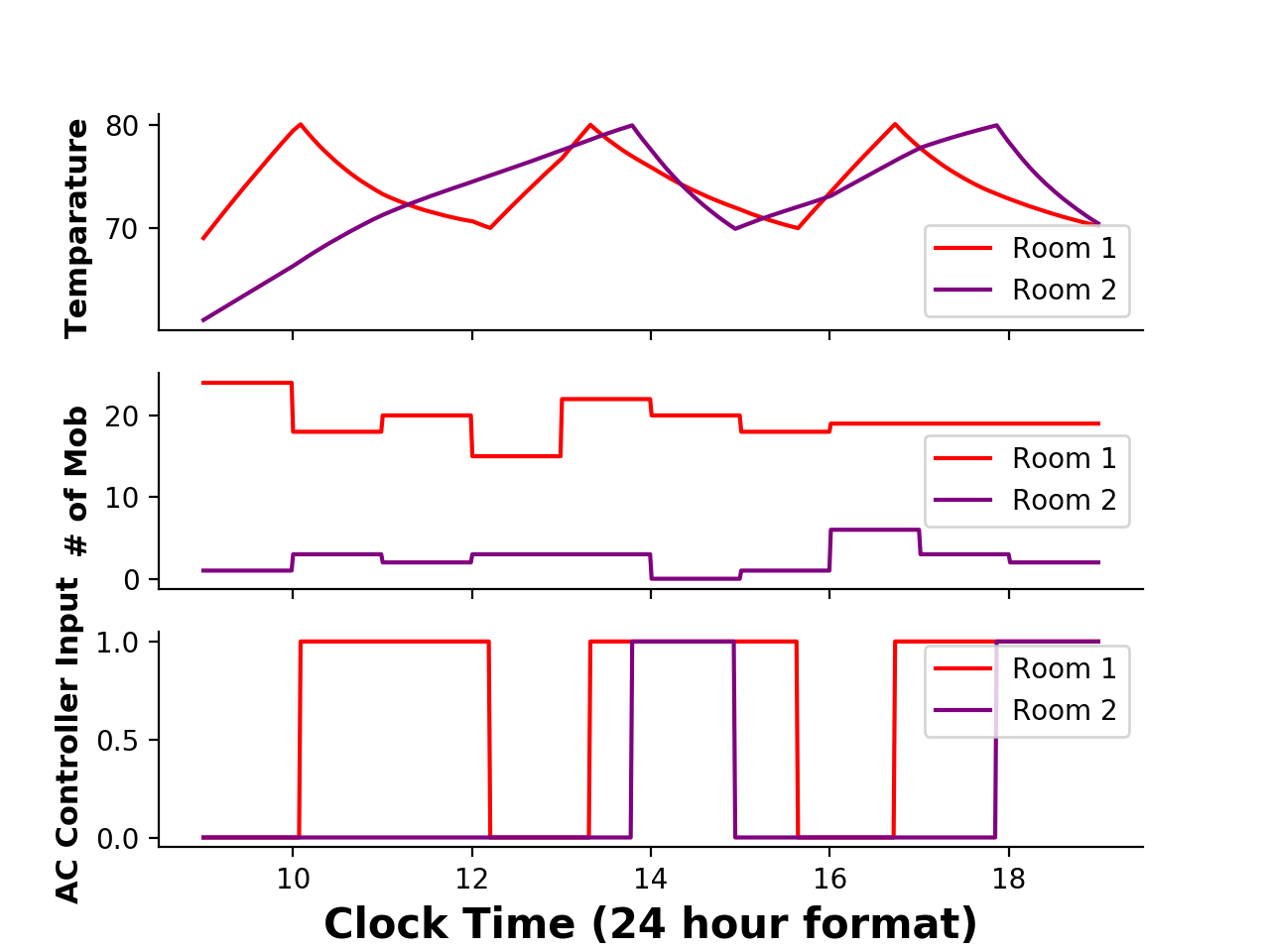}
    \caption{Big conference room (Room 1) and  Office Room's (Room 2) Sensor Temperature Reading from our simulation (above) and other simulation factors such as mob distribution (mid) and AC controller input(below - 1 means ON).}
    \label{fig:room1sensortempheattransfer}
\end{figure}

%
%
%
%
%
%
One thing to mention that we have performed 10 hours of
simulation. The time unit in our simulation will be in hour(h). Also we have run our simulator with all the temperature values in Kelvin($K$). For better understanding, we are mentioning the temperatures in Fahrenheit(F) in this paper. Finally, running our simulation, considering the parameters mentioned above and in \ref{subsec:supli1} and \ref{subsec:supli1}, the resulting outcomes following Eq. (\ref{eq:2}) can be viewed from Fig. \ref{fig:room1sensortempheattransfer}.

From Fig. \ref{fig:room1sensortempheattransfer}, we can see that the temperatures are remaining between the set points (70F and 80F). Similar kind of behavior was observed in Ref. \cite{privara2013building} and \cite{hazyuk2012optimal} with a slight variation as we have taken building occupants and some other extra factors into account. Therefore, we can be confident that the temperature data from our simulation is completely reasonable.

Putting this all together, we can rewrite Equation (\ref{eq:1}) as 

\begin{equation}\label{eq:2}
\frac{\partial T_i}{\partial t} = \sum_{j\sim i} \rho_{ij}(T_j(t)-T_i(t)) + \dot{Q}^{O} +  \sum_{l=1}^{n_i(t)}\qavg{l}\;.
\end{equation}

\section{Learning Formulation}\label{sec:3}
Now that we have simulation data of room temperatures in our hand, we can approach the main task of at hand, namely the following. If we are given the building and outside temperature data, together with the state of the HVAC, can we infer $n_i(t)$, the mobility of the people throughout the building, under mild assumptions?

\begin{figure*}[htp]
    \centering
    \includegraphics[width=0.85\textwidth,height=11cm]{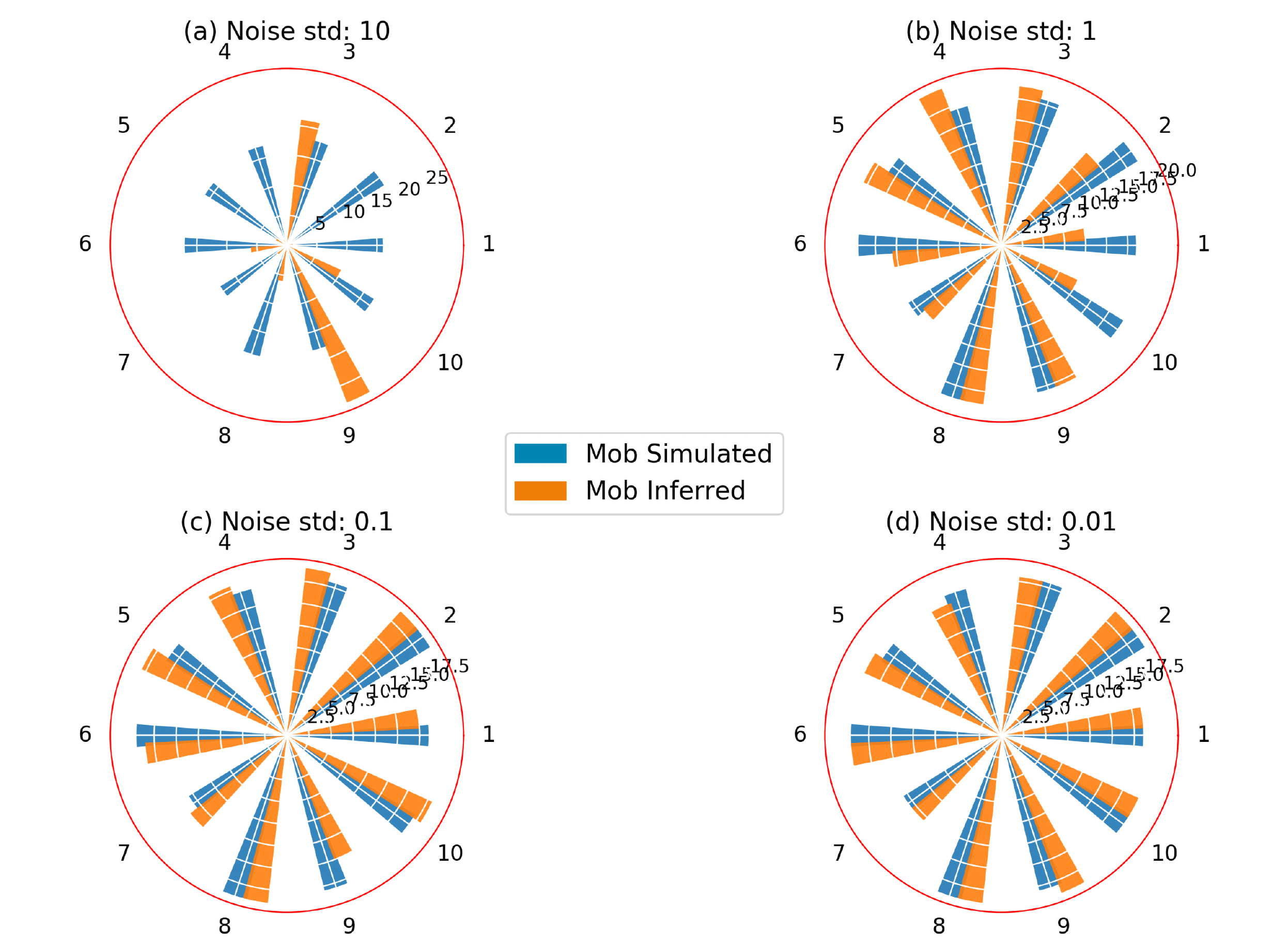}
    \caption{Mobility map of a large conference room from simulation (ground truth data) and inferred data from the reconstruction algorithm. Data displayed over 10 hours under different noise models. (a) Noise std, $\eta_{std}=10$  (b) $\eta_{std}=1$ (c) $\eta_{std}=0.1$ (d) $\eta_{std}=0.01$ 
}
    \label{fig:mobmapsunburstheattransfer}
\end{figure*}


In this section, we describe our \textbf{Reconstruction Algorithm (RA)} to recover the mobility $n_i(t)$. The Reconstruction Algorithm is based on a variation of the popular Least Square Estimator (LSE) with $L_2$ Regularization. The LSE gives us an estimate of $\sum_{l=1}^{n_i(t)}\qavg{l}$ from which we can infer $n_{i}(t)$.  The $L_2$ regularization ensures that the estimates are stable over short intervals of time. In the what follows, we explain the Reconstruction Algorithm in detail.

\textbf{Development of our method: } Clearly Equation (\ref{eq:2}) can be rewritten in matrix form to include each of the $k$ rooms in the building simultaneously. 

We assume the quantity $\rho_{H}$ is known, the $u_i(t)$ can be (perfectly) measured, and that the temperatures $T_i$ are measured with some noise. The quantities $\rho_I$ and $\rho_O$ are assumed to be unknown.  Finally, we assume the quantity $\hat{T}_{i}$ measured by the temperature sensors is includes independent noise sampled from a Guasian distriution:
\[
    \hat{T}_{i} = T_i + {\eta_i}\;.
\]
We now discretize the left hand side of Equation (\ref{eq:2}) to get
$$\mathbf{L}(\mathbf{T}):=\mathbf{L}=\frac{{\mathbf{T}}(t+1)-{\mathbf{T}}(t)}{\delta t}\;.$$
Let $\mathbf{D}$ be the diagonal matrix with $d_{ii}=\sum_{j\sim i} \rho_{ij}$.  Then the right hand side of Equation (\ref{eq:2}) can be written as
$$\mathbf{R}(\mathbf{T}):=\mathbf{R}=\bm{\rho}{\mathbf{T}}- \mathbf{D}{\mathbf{T}}+\dot{Q}^{O}\mathbf{1} + \dot{\mathbf{Q}}^{M}_i(t)\;.$$
Incorporating the noise from the temperature sensors, we have 
$\mathbf{L}(\hat{\mathbf{T}})-\mathbf{R}(\hat{\mathbf{T}})$ is a linear function of $\bm{\eta}$.
To infer the mobility map, $n_i(t)$, we also need to estimate the unknown model parameters.  Formally, we solve
\begin{equation}\label{eq:3}
\begin{aligned}
& \underset{\rho_I, \rho_O, \dot{Q}^O, \dot{\mathbf{Q}}^{M}_i(t) }{\text{argmin}}
& & \left\lVert \hat{\mathbf{L}}-\hat{\mathbf{R}} \right\rVert_{2}^{2} \\
& \text{s.t.} & &  \sum_{i=1}^{K}  \dot{Q}_i^{M}(t) = \sum_l \qavg{l} \\
& & & \dot{Q}_i^{M}(t)\geq 0, \;  \forall i,t \\
\end{aligned}
\end{equation}

Now we add a $L_2$ regularization term to (\ref{eq:3}) to force the mobility to be stable across consecutive time steps:

\begin{equation}\label{eq:4}
\begin{aligned}
& \underset{\mathbf{P}}{\text{argmin}}
& & \left\lVert \hat{\mathbf{L}}-\hat{\mathbf{R}} \right\rVert_{2}^{2}+\sum_{i=1}^{K}\sum_{t} \lambda \Bigg(\dot{Q}_i^{M}(t+1)- \dot{Q}_i^{M}(t)\Bigg)^2 \\
& & &   \\
\end{aligned}
\end{equation}
keeping the constraints the same as in (\ref{eq:3}) and optimizing over $\mathbf{P}= (\rho_I, \rho_O, \dot{Q}^O, \dot{\mathbf{Q}}^{M}_i(t))$. 

The choice of $\lambda$ will be discussed in Section \ref{sec:4}. Note that if there are $K$ rooms and $T$ time steps (for each room we collect the sensor data $T$ times at regular intervals of $\Delta t$) then there are $KT$ equations in this system and $KT+3$ parameters over which to optimize. Thus as $KT$ grows, we can expecct the LSE to yield better convex solutions. 

\textbf{Relaxing the constraints: } We focus on relaxing the first constraint in Eq. (\ref{eq:3}) which forces the total number of people in the building to be constant across time.  By relaxing this constraint, we can allow the number of people to fluctuate and also allow for the possibility that the exact number of people in the building is unknown.  To do this, let $\qavgavg$ be the average of the $\qavg{l}$ (or the expected value if these are unknown).
\begin{equation}\label{eq:5}
\begin{split}
N_{min}\qavgavg \leq\sum_{i=1}^{K}\dot{Q}^M_i(t)\leq N_{max} \qavgavg \\
(1-\epsilon_2)N\qavgavg \leq\sum_{i=1}^{K} \gamma \dot{Q}^M_i(t)\leq (1+\epsilon_1)N\qavgavg
\end{split}
\end{equation}
where, $\epsilon_1,\epsilon_2\geq0$. So, even if we specify a range of total mob, $[N_{min},N_{max}]$, we can still recover a reliable mobility map. Note that if $\epsilon_2>1$, we have ignored the $(1-\epsilon_2)N\qavgavg$ part because energy can not be negative. The effect of choosing $\epsilon_1$ and $\epsilon_2$ will be demonstrated in Section \ref{sec:4}.

We now have all the necessary ingredients to estimate the optimized parameter $\mathbf{\dot{Q}}^{M}(t)$.  The final step is to determine $n_{i}(t)$ by rounding:
\begin{equation}\label{eq:6}
\hat{n}_{i}(t):=\text{Round to Integer}\left(\frac{\hat{\dot{Q}}^{M}_i(t)}{ \qavgavg}\right)
\end{equation}

It can easily be seen that the performance of our Reconstruction Algorithm depends on $\epsilon, \lambda \text{ and } \eta$ whose effects will be discussed in Section \ref{sec:4}.


\section{\label{sec:4}Results}

Now that we have described our Simulation, Problem Formulation and our Reconstruction Algorithm (\textbf{RA}), in this section, we show-case the output and performance of RA as well as the effect of various model parameters ($\epsilon, \lambda \text{ and } \eta$). 

\textbf{Output of RA applied on simulated temperature data: }
If we apply RA on the simulated data using Eq. (\ref{eq:4}) and (\ref{eq:6}), we can successfully recover the  model parameters, $\routside, \rinside, \otherheat, \dot{Q}_i^{M}(t)$ and finally obtain the estimate $\hat{n}_i(t)$. We have recovered:

$$\routside = 0.099  \text{ , } \rinside = 0.101 \text{ and } \otherheat = 1.35$$ which is almost exact to the parameters used in simulation.

\begin{figure}[t]
    \centering
    \includegraphics[width=9cm]{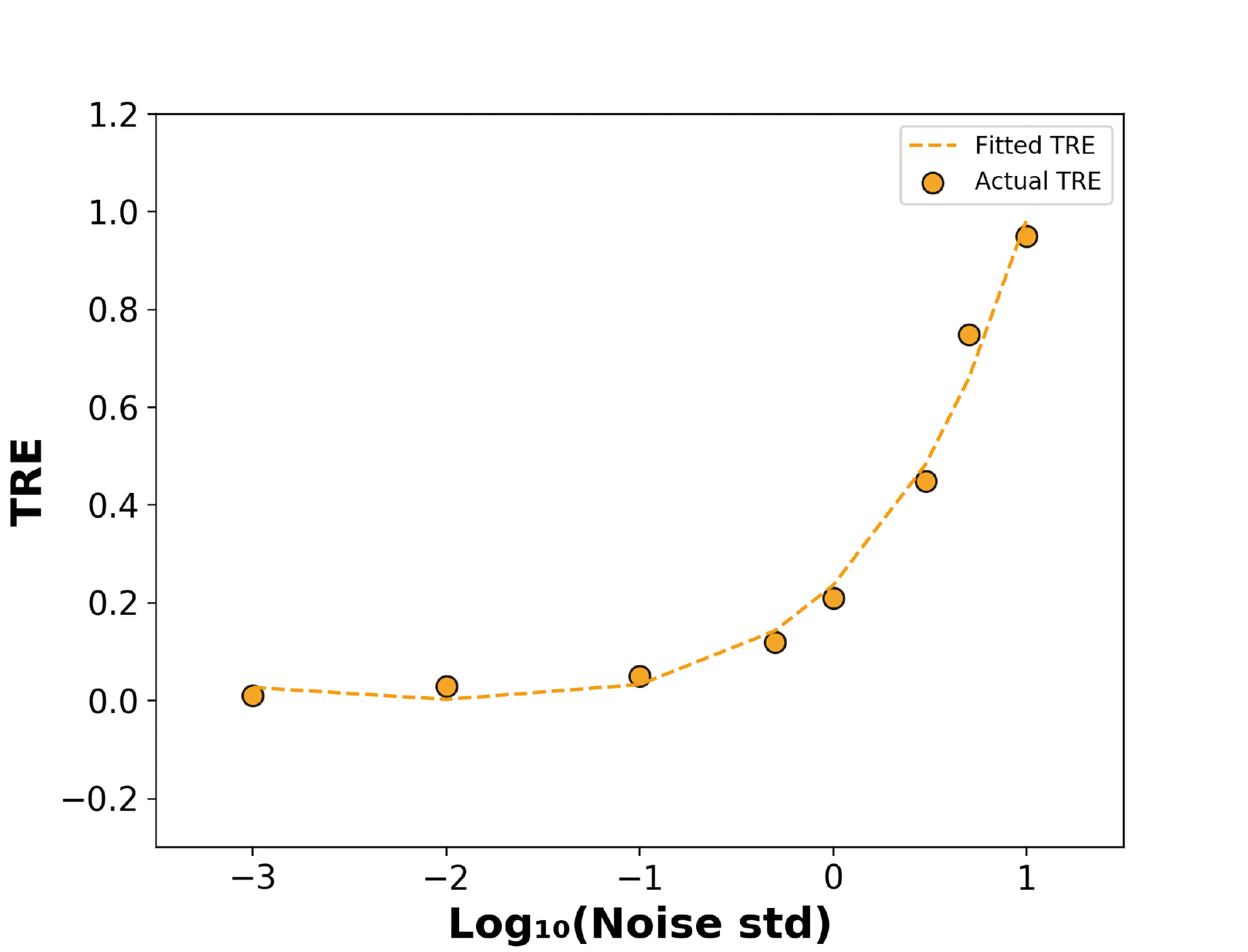}
    \caption{TRE of RA vs $Log_{10}$ of $\eta_{std}$ [K=16, N= 45, $\Delta t=1$ minute, $T=600$ minutes, $\lambda=0.1$, $\epsilon=0.2$].  }
    \label{fig:mobmapNMAEheattransfer}
\end{figure}

Fig. \ref{fig:mobmapsunburstheattransfer} demonstrates our constructed mobility map, $\hat{n}_i(t)$, using Eq. (\ref{eq:6}) with the variation of noise. Note that our estimates are therefor updated in every time step (here one minute): in our reconstruction we took the median number of people returned by the reconstruction algorithm as the best approximation for that hour (see Figure \ref{fig:mobmapsunburstheattransfer}). When we implemented the reconstruciton algorithm on a real data set, we took the best approximation over thirty minute windows and achieved reasonable results (shown below in this Section). In practice the appropriate value for this time frame could vary depending on the situation and application. 

Note that, as shown in Figure \ref{fig:mobmapsunburstheattransfer}, that when the standard deviation of the noise is high ($\eta_{std}=10$ in panel (a)), the reconstruction algorithm performs poorly. On the other hand, for the smaller values of  $\eta_{std}$ shown in panels  (b), (c) and (d), the inference reasonably matchs the number of people in the room from the simulation. 

\textbf{Performace evaluation of RA: }
We measure the inference performance of RA by a term called \textit{Total Reconstruction Error \textbf{(TRE)}} which is a function of \textit{Normalized Mean Absolute Error \textbf{(NMAE)}}. The lower the TRE is, the better the performance is. We define TRE and NMAE as follows:

\[
    \text{TRE} = \frac{\sum_{i}^{k} \text{NMAE}_i}{k}\\
\]
\[
    \text{NMAE}_i = \frac{\sum_{t=1}^{T}|n_i(t) - \hat{n}_i(t)|}{\sum_{t=1}^T n_i(t)}\\
\]

We measure the performance of RA on the basis of TRE with respect to the variation of the model parameters (i)$\eta_{std}$ ,(ii)$\lambda$ and (iii) $\epsilon$.

\textbf{(i) Effect of $\eta_{std}$ on TRE: }
Fig. \ref{fig:mobmapNMAEheattransfer} exhibits the effect of  $\eta_{std}$ (actually $Log_{10}$ of $\eta_{std}$)  on the performance of RA in terms of TRE. When $\eta_{std}\leq 1$ ($\leq $0 along X axis), we get TRE $\leq 0.2$ which is similar to having maximum 20\% error in reconstruction. If we consider that as our tolerance level, we can allow maximum Gaussian noise with standard deviation 1. Beyond that standard deviation, the reconstruction performance will degrade largely. Therefore, we recommend the level of accepted $\eta_{std}$ in range [0,1] to have a reliable Mob Map. 
\begin{figure}[t]
    \centering
    \includegraphics[width=9cm]{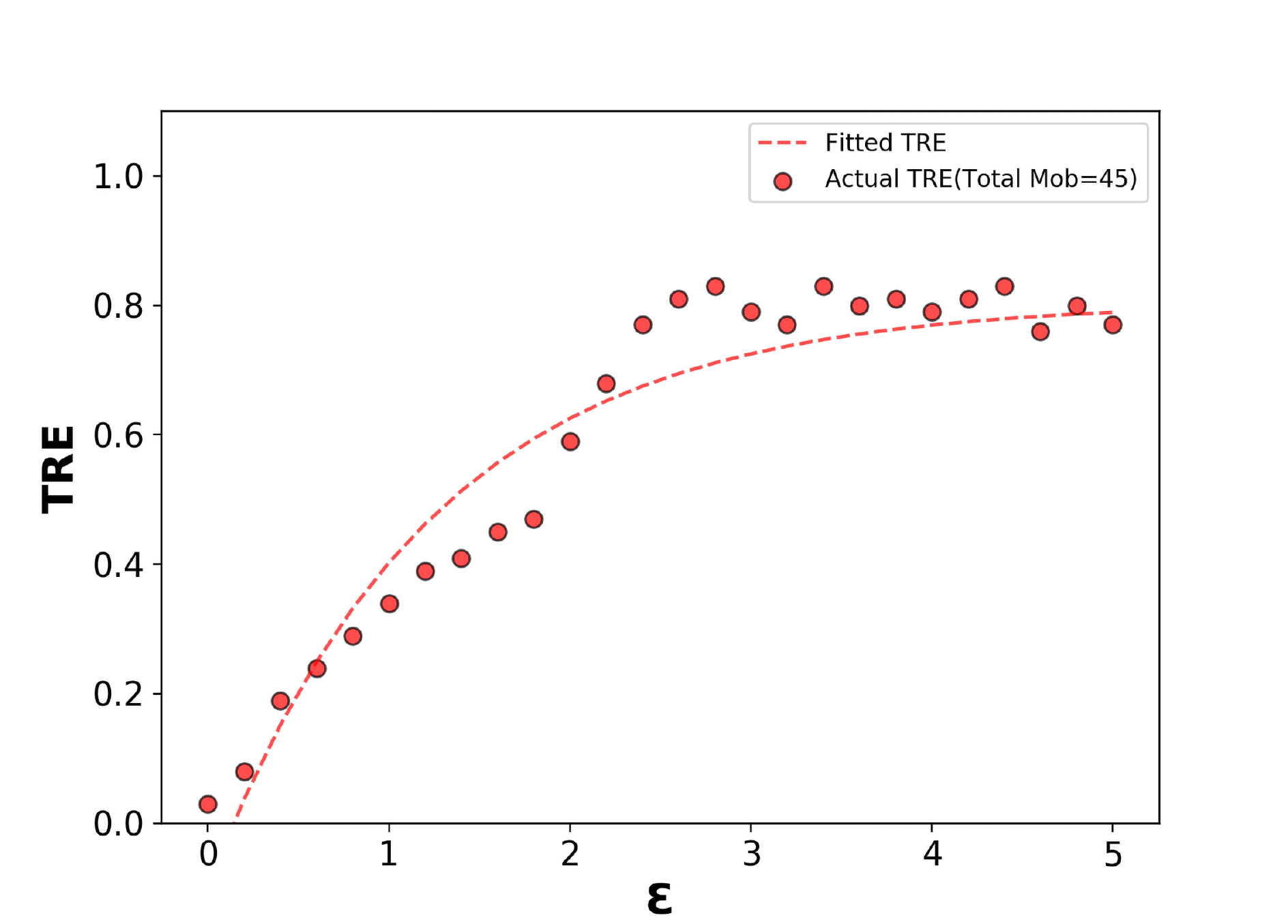}
    \caption{TRE of RA vs $\epsilon$ [K=16, N= 45, $\Delta t=1 minute$, $T=600$ minutes, $\lambda=0.1$, $\eta_{std}=0.1$].}
    \label{fig:epsilonvstre}
\end{figure}

\textbf{(ii) Effect of $\epsilon$ on TRE: }In Section \ref{sec:3}, we mentioned that our inference performance varies depending on the choice of $\epsilon_1$ and $\epsilon_2$ in constraint (\ref{eq:5}). We will show the effect of choosing $\epsilon_1$ and $\epsilon_2$ by considering $\epsilon_1=\epsilon_2=\epsilon$.
In Figure \ref{fig:epsilonvstre}, we have shown the effect of $\epsilon$ on the TRE. It is evident from Figure \ref{fig:epsilonvstre} that choosing a smaller (more accurate) range of $[N_{min},N_{max}](\epsilon \approx 0)$, yields better mobility map reconstructions (i.e. lower TRE).  Setting $\epsilon=0$ is equivalent to using the original constraint from (\ref{eq:3}) which is appropriate when the exact number of people in the building is kown (and constant). Setting $\epsilon>2$ is actually equivalent to using no constraint at all. As usual, the more one knows, the easier inference is, or int this case, the smaller  $\epsilon$, the better the performance of the reconstruction algorithm will be.

\begin{figure}[t]
    \centering
    \includegraphics[width=9cm]{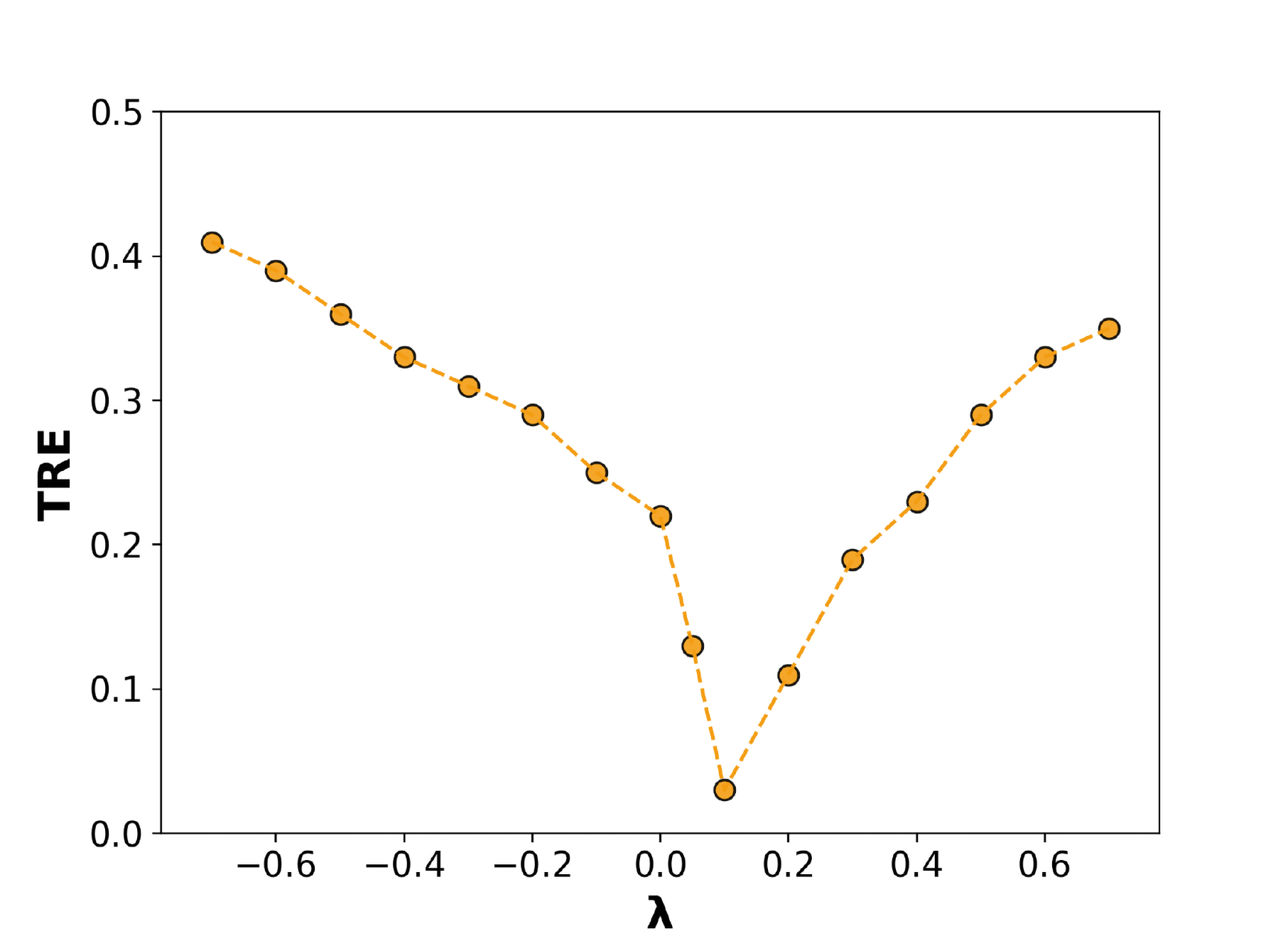}
    \caption{TRE of RA vs $\lambda$ [K=16, N= 45, $\Delta t=1$ minute, $T=600$ minutes, $\epsilon=0.2$, $\eta_{std}=0.1$].}
    \label{fig:lambdavstre}
\end{figure}

\textbf{(iii) Effect of $\lambda$ on TRE: }The last thing we want to discuss is the effect of $\lambda$ on map reconstruction performance. From the empirical result shown in Fig. \ref{fig:lambdavstre}, we see that when $\lambda=0.1$, we get the lowest TRE, meaning better reconstruction. We recommend to use  $\lambda=0.1$ but depending on the recommended max tolerance level (20\%), $\lambda$ can have a value in range $\lambda=[0.03,0.3]$

\textbf{Implementation on Real Data set:}
As we have mentioned earlier, no data set is complete to run our method. That is why we choose to simulate the room temperatures. Till now we have seen the performance of RA on the simulated data. However, to show the effectiveness of RA in different scenario, we are implementing our method of RA to some real toy data sets. We assume reasonable values for the missing arguments for our Algorithm.

\begin{figure}[t]
    \centering
    \includegraphics[width=9cm]{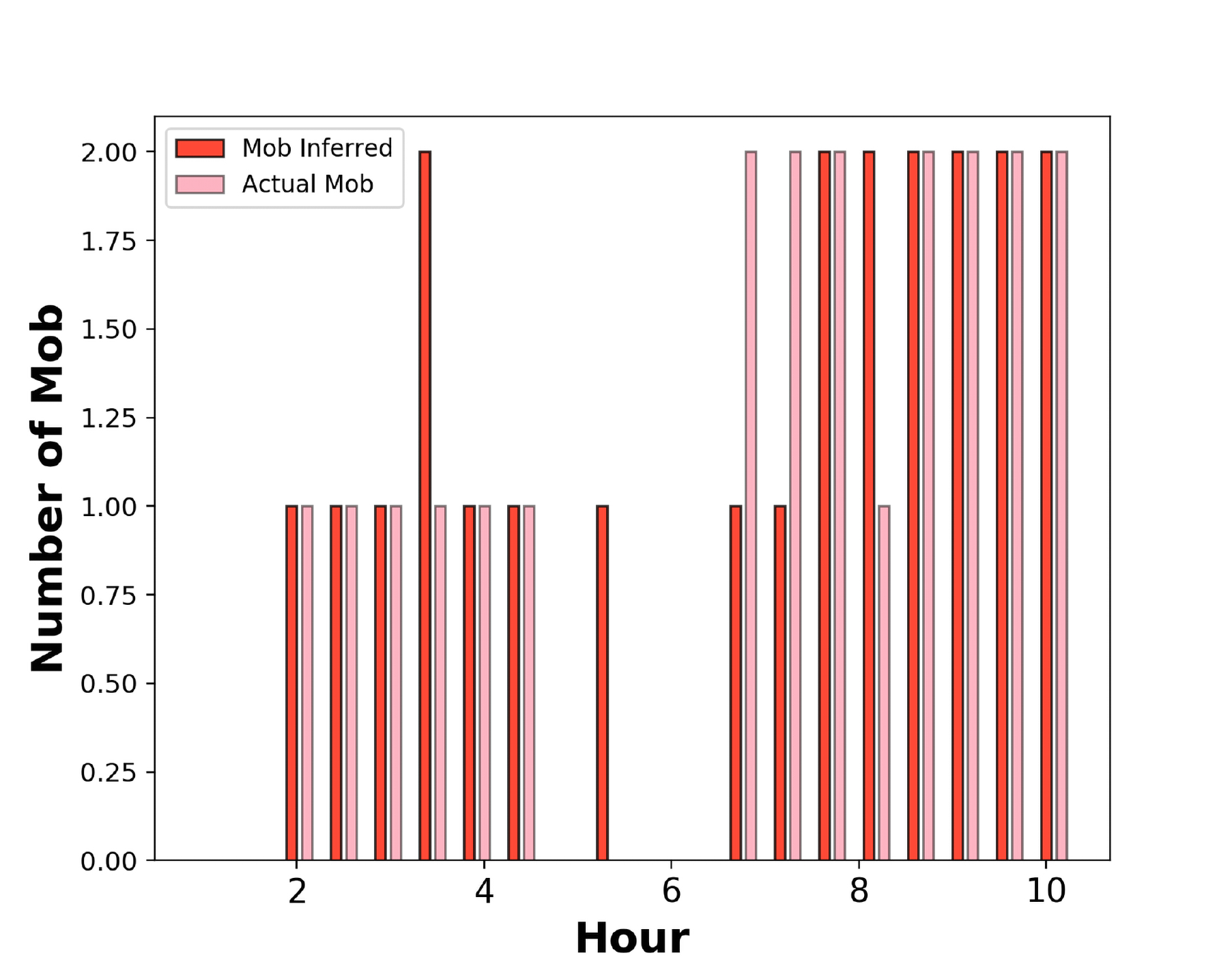} 
    \caption{Mobility Map generated from Data set 1 ($TR=30$ minutes). At each TR, left column is our inferred mob number and right column is the actual mob number.[$\Delta t=3$ minutes, $T=200$ minutes, $\epsilon=0.2$, $\lambda=0.1$]}
    \label{fig:germany}
\end{figure}

\textbf{(i) Data set 1:} This data, in \cite{armknecht2019privacy}, was collected from March to April, 2016. We have filtered the data set to get 10 hours of data(sampled every 3 minutes) from April 28, 2016 . It contains the temperature of only one room as well as the number of occupants in that room. As a consequence, we get ground truth to compare our results. The only problem of this data set is that we do not have some other distributions like ambient temperatures and HVAC status (ON/OFF or cool air temperature). We assume reasonable values for these. For example, we set $T^{O}(t)$ from a distribution $(\mu=55F, \sigma=2)$ which is compatible with ambient temperature in Mannheim, Germany from where the data was collected. Another thing is that because of the average ambient temperature being very cool($\mu=55F$), we assume that the air conditioner is OFF. Considering these situations in mind, we have applied RA on this data set and got the Mob Map like in Fig. \ref{fig:germany}. Moreover, the relative higher value of $\otherheat = 2.31$ (in our simulation $\otherheat \text{ was } 1.36$) delivered by RA indicates the presence of possible Heating System inside the room. 

\textbf{(ii) Data set 2:} We have filtered this data set in \cite{candanedo2017data} to collect 10 hours (9:00 to 19:00) of data sampled in every 10 minutes from May 26, 2016. The room architecture (neighbour room information) is known and we have gathered the data of all four rooms in lower floor. However, we can get $T^{O}(t)$ and $T_i(t)$ directly from the data set. To get an idea of AC's ON/OFF status, we have observed that, for Room 3, the temperature dropped abruptly from 11:00 to 12:30 and from 14:00 to 17:00. We have assumed that in this time the AC was ON. Another thing is that there was no hints given about HVAC's air volume capacity or temperature of the cooled air. For this, we have used the values of $\rhvac$ and $T^{H}$ from our simulation. 
\begin{figure}[htp]
    \centering
    \includegraphics[width=9cm]{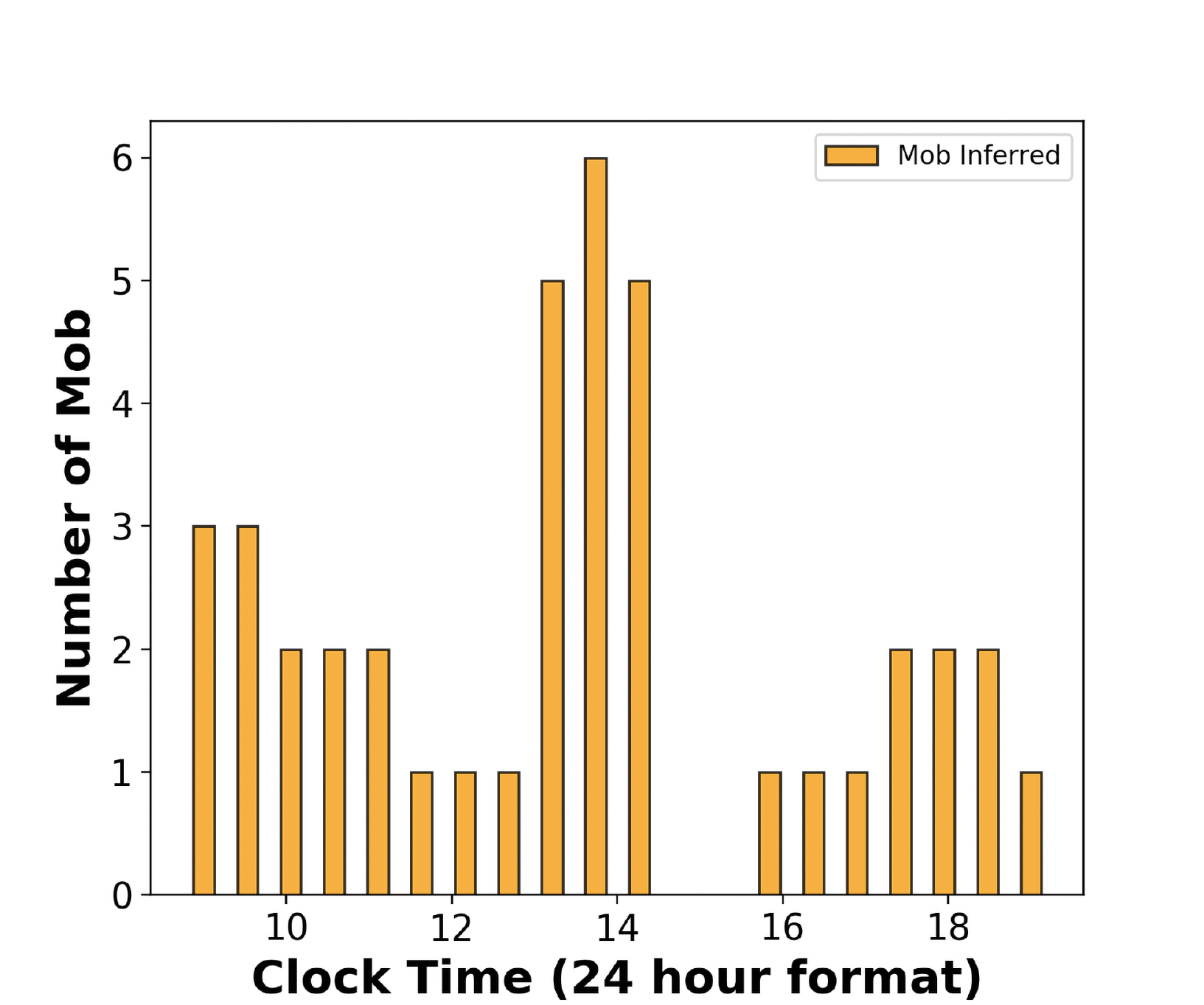} 
    \caption{Moblity Map generated from Data set 2 ($TR=30$ minutes).[$\Delta t=3$ minutes, $T=200$ minutes, $\epsilon=0.2$, $\lambda=0.1$]}
    \label{fig:house}
\end{figure}
Considering these assumptions, we have run RA on this data set and got the Mob Map of Room 3 similar to Fig. \ref{fig:house}. Note that there is no ground truth on the number of occupant in this data set. We have chosen Room 3 to show the Mob Map here because this room is attached with Room 1 and Room 2 as well as with the environment which is almost similar to our simulation design.

\section{\label{sec:5}Discussion and Conclusion}
In this paper, we simulated a thermal modeling of a building and proposed a method to infer the room occupancies over time from easily available data (standard HVAC data, the building layout, and external temperature data).   We demonstrate that occupancy levels can be reconstructed under realistic noise assumptions.  We also demonstrate the efficacy and performance of the reconstruction algorithm with respect to model parameters such as the noise level from sensor data, choice of occupancy constraints and other model parameters.  We illustrate the effectiveness of the reconstruction algorithm by testing it on synthetic data as well as on two publicly available data sets from the real world. 

Unfortunately, while the inputs to our algorithm are easily collected, finding labeled data sets (with ground-truth occupancy levels) has been a challenge. In future work, we hope to apply our method on a real-world data set with such ground truth labels to further verify our method. Finally, we mention that though our simulation model is moderately complex, there is room to construct more complex and realistic thermal models, by including effects such as solar radiation and heat transfer via convection.


\section{\label{sec:6}Supplementary Materials}
\subsection{Choice of the Constants:}\label{subsec:supli1}
\textbf{$C_i$}: $C_i$ is the Heat Capacity for entire room. All the elements (air, furniture and walls) in a room contribute to it.
$$C_i= C_{air} + C_{furniture} + C_{wall} = (C_{volumetric}* volume) +$$ 
$$+(\text{Furniture mass * Specific Heat Capacity,}C_p ) + $$
$$+(\rho_{wall}*volume_{wall}*(C_p)_{wall})$$
Considering standard values, we get,
$$C_i=(1.29KJm^{-3}K^{-1} * 30m^3) + (4200 J/kgK * 200kg) +$$ $$+(2.13*10^5 gm^{-3}*5m^3*0.84 J/gK) = 1*10^6 J/K$$
$a_{ext}$: Every room performs heat transfer with the ambient temperature via conduction process in the wall and window glass. The equation for $a_{ext}$ is:
$$a_{ext} = h_{wall}*A_{wall} + h_{glass}*A_{glass}$$
where, h is the Thermal Transmittance of a substance. Using standard values we get:
$$a_{ext} = (1Wm^{-2}K^{-1} * 20m^2)+(5Wm^{-2}K^{-1} * 2m^2) = 30W/K$$

$a_{room}$: Every room performs heat transfer with neighbouring rooms via conduction process in the wall only. The equation is same as $a_{ext}$ but excluding the glass part. Using standard values we get:
$$a_{rooms} = (1Wm^{-2}K^{-1} * 20m^2) = 20W/K$$\newline

$\otherheat$: Considering 3 computers (100W energy consumed per computer), 4 lights (10 W per light) and 1 printer(40W/printer) per room, each room will radiate a total of 380W = $1.36*10^6$J/h energy.\newline

$\qavgavg$: Average heat produced by an average human body while staying indoor is given by the equation:
$$\qavgavg = M*A_{body}$$
Considering standard Metabolic rate, $M = 55Wm^{-2}$ and average body area, $A_{body} = 2 m^2$ we get, $\qavgavg = 110W = 3.96*10^5 J/h$\newline

$\rho \dot{V}_{HVAC} C_p$: Considering each room 250 sqr-ft and standard rules of 1 Cubic ft per minute (1 CFM) per sqr-ft, we get $\dot{V}_{HVAC} = 250 CFM = 450 m^3/h$. Using usual values, we get $\rho \dot{V}_{HVAC} C_p$ = $\rho_{air} \dot{V}_{HVAC} (C_p)_{air}= 1.225 kgm^{-3} * 450 m^3/h *  1000 J/kgK = 5.51*10^5 J/hK$

Now,
$\routside = \frac{a_{ext}}{C_i} = \frac{30W/K}{10^6 J/K}= 0.1h^{-1}$

$\rinside = \frac{a_{rooms}}{C_i} = \frac{20W/K}{10^6 J/K}= 0.1h^{-1}$

$\otherheat = \frac{\dot{Q}_{internal}}{C_i} = \frac{1.36*10^6J/h}{10^6 J/K}=1.36 Kh^{-1}$


$\rhvac = \frac{\rho \dot{V}_{HVAC} C_p}{C_i} = \frac{5.51*10^5 J/hK}{10^6 J/K}= 0.6h^{-1}$\newline

\subsection{Simulation Parameters:}\label{subsec:supli2}

$\routside = 0.1h^{-1}$, $\rinside = 0.1h^{-1}$, $\otherheat =1.36 Kh^{-1}$, and 
$\rhvac = 0.6h^{-1}$

$[T_i^{min},T_i^{max}] = [70F, 80F]$

Total number of rooms, $k$ = 16 and Total mob in the building, $N$ = 45 [Distributed over $k$ rooms]

Number of Big Conference room, BCR = 1; number of Small Conference room, SCR = 1; number of Normal Office room, OR = 14.

Weight for Big conference room, $w_{i}^{b}$ = 5; weight for small conference room, $w_{i}^{s}$ = 3; weight for office room, $w_{i}^{o}$ = 1

Total time for simulation = 10 hours; time step for solving the equation, $\Delta t$ = 10/600 hour = 1 minute; Time range, $TR=1$ hour.

\subsection{Implementation Platform Details:}\label{subsec:supli3}
We have used Python platform (Python 3.7.2) and SQSLP optimization algorithm from  scipy package in order to manipulate the LSE and $L_2$ Regularization to  optimize the model parameters. All the codes were run in Macbook Pro, 2017, 2.3 GHz Intel Core i5, 16 GB memory.

\bibliography{arxivMobility}

\providecommand{\noopsort}[1]{}\providecommand{\singleletter}[1]{#1}%
\begin{thebibliography}{47}%
\makeatletter
\providecommand \@ifxundefined [1]{%
 \@ifx{#1\undefined}
}%
\providecommand \@ifnum [1]{%
 \ifnum #1\expandafter \@firstoftwo
 \else \expandafter \@secondoftwo
 \fi
}%
\providecommand \@ifx [1]{%
 \ifx #1\expandafter \@firstoftwo
 \else \expandafter \@secondoftwo
 \fi
}%
\providecommand \natexlab [1]{#1}%
\providecommand \enquote  [1]{``#1''}%
\providecommand \bibnamefont  [1]{#1}%
\providecommand \bibfnamefont [1]{#1}%
\providecommand \citenamefont [1]{#1}%
\providecommand \href@noop [0]{\@secondoftwo}%
\providecommand \href [0]{\begingroup \@sanitize@url \@href}%
\providecommand \@href[1]{\@@startlink{#1}\@@href}%
\providecommand \@@href[1]{\endgroup#1\@@endlink}%
\providecommand \@sanitize@url [0]{\catcode `\\12\catcode `\$12\catcode
  `\&12\catcode `\#12\catcode `\^12\catcode `\_12\catcode `\%12\relax}%
\providecommand \@@startlink[1]{}%
\providecommand \@@endlink[0]{}%
\providecommand \url  [0]{\begingroup\@sanitize@url \@url }%
\providecommand \@url [1]{\endgroup\@href {#1}{\urlprefix }}%
\providecommand \urlprefix  [0]{URL }%
\providecommand \Eprint [0]{\href }%
\providecommand \doibase [0]{https://doi.org/}%
\providecommand \selectlanguage [0]{\@gobble}%
\providecommand \bibinfo  [0]{\@secondoftwo}%
\providecommand \bibfield  [0]{\@secondoftwo}%
\providecommand \translation [1]{[#1]}%
\providecommand \BibitemOpen [0]{}%
\providecommand \bibitemStop [0]{}%
\providecommand \bibitemNoStop [0]{.\EOS\space}%
\providecommand \EOS [0]{\spacefactor3000\relax}%
\providecommand \BibitemShut  [1]{\csname bibitem#1\endcsname}%
\let\auto@bib@innerbib\@empty
\bibitem [{\citenamefont {Faruque}\ \emph {et~al.}(2016)\citenamefont
  {Faruque}, \citenamefont {Abdullah}, \citenamefont {Chhetri}, \citenamefont
  {Canedo},\ and\ \citenamefont {Wan}}]{faruque2016acoustic}%
  \BibitemOpen
  \bibfield  {author} {\bibinfo {author} {\bibfnamefont {A.}~\bibnamefont
  {Faruque}}, \bibinfo {author} {\bibfnamefont {M.}~\bibnamefont {Abdullah}},
  \bibinfo {author} {\bibfnamefont {S.~R.}\ \bibnamefont {Chhetri}}, \bibinfo
  {author} {\bibfnamefont {A.}~\bibnamefont {Canedo}},\ and\ \bibinfo {author}
  {\bibfnamefont {J.}~\bibnamefont {Wan}},\ }\bibfield  {title} {\bibinfo
  {title} {Acoustic side-channel attacks on additive manufacturing systems},\
  }in\ \href@noop {} {\emph {\bibinfo {booktitle} {Proceedings of the 7th
  International Conference on Cyber-Physical Systems}}}\ (\bibinfo
  {organization} {IEEE Press},\ \bibinfo {year} {2016})\ p.~\bibinfo {pages}
  {19}\BibitemShut {NoStop}%
\bibitem [{\citenamefont {Kac}(1966)}]{kac1966can}%
  \BibitemOpen
  \bibfield  {author} {\bibinfo {author} {\bibfnamefont {M.}~\bibnamefont
  {Kac}},\ }\bibfield  {title} {\bibinfo {title} {Can one hear the shape of a
  drum?},\ }\href@noop {} {\bibfield  {journal} {\bibinfo  {journal} {The
  american mathematical monthly}\ }\textbf {\bibinfo {volume} {73}},\ \bibinfo
  {pages} {1} (\bibinfo {year} {1966})}\BibitemShut {NoStop}%
\bibitem [{\citenamefont {Zhou}\ \emph {et~al.}(2014)\citenamefont {Zhou},
  \citenamefont {Tian}, \citenamefont {Xu}, \citenamefont {Yu}, \citenamefont
  {Hong},\ and\ \citenamefont {Wu}}]{zhou2014scanme}%
  \BibitemOpen
  \bibfield  {author} {\bibinfo {author} {\bibfnamefont {M.}~\bibnamefont
  {Zhou}}, \bibinfo {author} {\bibfnamefont {Z.}~\bibnamefont {Tian}}, \bibinfo
  {author} {\bibfnamefont {K.}~\bibnamefont {Xu}}, \bibinfo {author}
  {\bibfnamefont {X.}~\bibnamefont {Yu}}, \bibinfo {author} {\bibfnamefont
  {X.}~\bibnamefont {Hong}},\ and\ \bibinfo {author} {\bibfnamefont
  {H.}~\bibnamefont {Wu}},\ }\bibfield  {title} {\bibinfo {title} {Scanme:
  location tracking system in large-scale campus wi-fi environment using
  unlabeled mobility map},\ }\href@noop {} {\bibfield  {journal} {\bibinfo
  {journal} {Expert Systems with Applications}\ }\textbf {\bibinfo {volume}
  {41}},\ \bibinfo {pages} {3429} (\bibinfo {year} {2014})}\BibitemShut
  {NoStop}%
\bibitem [{\citenamefont {Horng}\ \emph {et~al.}(2011)\citenamefont {Horng},
  \citenamefont {Chen}, \citenamefont {Ferng}, \citenamefont {Kao},\ and\
  \citenamefont {Li}}]{horng2011enhancing}%
  \BibitemOpen
  \bibfield  {author} {\bibinfo {author} {\bibfnamefont {S.-J.}\ \bibnamefont
  {Horng}}, \bibinfo {author} {\bibfnamefont {C.}~\bibnamefont {Chen}},
  \bibinfo {author} {\bibfnamefont {H.-W.}\ \bibnamefont {Ferng}}, \bibinfo
  {author} {\bibfnamefont {T.-W.}\ \bibnamefont {Kao}},\ and\ \bibinfo {author}
  {\bibfnamefont {M.-H.}\ \bibnamefont {Li}},\ }\bibfield  {title} {\bibinfo
  {title} {Enhancing wlan location privacy using mobile behavior},\ }\href@noop
  {} {\bibfield  {journal} {\bibinfo  {journal} {Expert Systems with
  Applications}\ }\textbf {\bibinfo {volume} {38}},\ \bibinfo {pages} {175}
  (\bibinfo {year} {2011})}\BibitemShut {NoStop}%
\bibitem [{\citenamefont {Feng}\ \emph {et~al.}(2020)\citenamefont {Feng},
  \citenamefont {Mehmani},\ and\ \citenamefont {Zhang}}]{feng2020deep}%
  \BibitemOpen
  \bibfield  {author} {\bibinfo {author} {\bibfnamefont {C.}~\bibnamefont
  {Feng}}, \bibinfo {author} {\bibfnamefont {A.}~\bibnamefont {Mehmani}},\ and\
  \bibinfo {author} {\bibfnamefont {J.}~\bibnamefont {Zhang}},\ }\bibfield
  {title} {\bibinfo {title} {Deep learning-based real-time building occupancy
  detection using ami data},\ }\href@noop {} {\bibfield  {journal} {\bibinfo
  {journal} {IEEE Transactions on Smart Grid}\ } (\bibinfo {year}
  {2020})}\BibitemShut {NoStop}%
\bibitem [{\citenamefont {Razavi}\ \emph {et~al.}(2019)\citenamefont {Razavi},
  \citenamefont {Gharipour}, \citenamefont {Fleury},\ and\ \citenamefont
  {Akpan}}]{razavi2019occupancy}%
  \BibitemOpen
  \bibfield  {author} {\bibinfo {author} {\bibfnamefont {R.}~\bibnamefont
  {Razavi}}, \bibinfo {author} {\bibfnamefont {A.}~\bibnamefont {Gharipour}},
  \bibinfo {author} {\bibfnamefont {M.}~\bibnamefont {Fleury}},\ and\ \bibinfo
  {author} {\bibfnamefont {I.~J.}\ \bibnamefont {Akpan}},\ }\bibfield  {title}
  {\bibinfo {title} {Occupancy detection of residential buildings using smart
  meter data: A large-scale study},\ }\href@noop {} {\bibfield  {journal}
  {\bibinfo  {journal} {Energy and Buildings}\ }\textbf {\bibinfo {volume}
  {183}},\ \bibinfo {pages} {195} (\bibinfo {year} {2019})}\BibitemShut
  {NoStop}%
\bibitem [{\citenamefont {Armknecht}\ \emph {et~al.}(2019)\citenamefont
  {Armknecht}, \citenamefont {Benenson}, \citenamefont {Morgner}, \citenamefont
  {M{\"u}ller},\ and\ \citenamefont {Riess}}]{armknecht2019privacy}%
  \BibitemOpen
  \bibfield  {author} {\bibinfo {author} {\bibfnamefont {F.}~\bibnamefont
  {Armknecht}}, \bibinfo {author} {\bibfnamefont {Z.}~\bibnamefont {Benenson}},
  \bibinfo {author} {\bibfnamefont {P.}~\bibnamefont {Morgner}}, \bibinfo
  {author} {\bibfnamefont {C.}~\bibnamefont {M{\"u}ller}},\ and\ \bibinfo
  {author} {\bibfnamefont {C.}~\bibnamefont {Riess}},\ }\bibfield  {title}
  {\bibinfo {title} {Privacy implications of room climate data},\ }\href@noop
  {} {\bibfield  {journal} {\bibinfo  {journal} {Journal of Computer Security}\
  }\textbf {\bibinfo {volume} {27}},\ \bibinfo {pages} {113} (\bibinfo {year}
  {2019})}\BibitemShut {NoStop}%
\bibitem [{\citenamefont {Golestan}\ \emph {et~al.}(2018)\citenamefont
  {Golestan}, \citenamefont {Kazemian},\ and\ \citenamefont
  {Ardakanian}}]{golestan2018data}%
  \BibitemOpen
  \bibfield  {author} {\bibinfo {author} {\bibfnamefont {S.}~\bibnamefont
  {Golestan}}, \bibinfo {author} {\bibfnamefont {S.}~\bibnamefont {Kazemian}},\
  and\ \bibinfo {author} {\bibfnamefont {O.}~\bibnamefont {Ardakanian}},\
  }\bibfield  {title} {\bibinfo {title} {Data-driven models for building
  occupancy estimation},\ }in\ \href@noop {} {\emph {\bibinfo {booktitle}
  {Proceedings of the Ninth International Conference on Future Energy
  Systems}}}\ (\bibinfo {year} {2018})\ pp.\ \bibinfo {pages}
  {277--281}\BibitemShut {NoStop}%
\bibitem [{\citenamefont {Ardakanian}\ \emph {et~al.}(2016)\citenamefont
  {Ardakanian}, \citenamefont {Bhattacharya},\ and\ \citenamefont
  {Culler}}]{ardakanian2016non}%
  \BibitemOpen
  \bibfield  {author} {\bibinfo {author} {\bibfnamefont {O.}~\bibnamefont
  {Ardakanian}}, \bibinfo {author} {\bibfnamefont {A.}~\bibnamefont
  {Bhattacharya}},\ and\ \bibinfo {author} {\bibfnamefont {D.}~\bibnamefont
  {Culler}},\ }\bibfield  {title} {\bibinfo {title} {Non-intrusive techniques
  for establishing occupancy related energy savings in commercial buildings},\
  }in\ \href@noop {} {\emph {\bibinfo {booktitle} {Proceedings of the 3rd ACM
  International Conference on Systems for Energy-Efficient Built
  Environments}}}\ (\bibinfo {year} {2016})\ pp.\ \bibinfo {pages}
  {21--30}\BibitemShut {NoStop}%
\bibitem [{\citenamefont {Arief-Ang}\ \emph
  {et~al.}(2017{\natexlab{a}})\citenamefont {Arief-Ang}, \citenamefont
  {Salim},\ and\ \citenamefont {Hamilton}}]{arief2017sd}%
  \BibitemOpen
  \bibfield  {author} {\bibinfo {author} {\bibfnamefont {I.~B.}\ \bibnamefont
  {Arief-Ang}}, \bibinfo {author} {\bibfnamefont {F.~D.}\ \bibnamefont
  {Salim}},\ and\ \bibinfo {author} {\bibfnamefont {M.}~\bibnamefont
  {Hamilton}},\ }\bibfield  {title} {\bibinfo {title} {Sd-hoc: Seasonal
  decomposition algorithm for mining lagged time series},\ }in\ \href@noop {}
  {\emph {\bibinfo {booktitle} {Australasian Conference on Data Mining}}}\
  (\bibinfo {organization} {Springer},\ \bibinfo {year} {2017})\ pp.\ \bibinfo
  {pages} {125--143}\BibitemShut {NoStop}%
\bibitem [{\citenamefont {Ghai}\ \emph {et~al.}(2012)\citenamefont {Ghai},
  \citenamefont {Thanayankizil}, \citenamefont {Seetharam},\ and\ \citenamefont
  {Chakraborty}}]{ghai2012occupancy}%
  \BibitemOpen
  \bibfield  {author} {\bibinfo {author} {\bibfnamefont {S.~K.}\ \bibnamefont
  {Ghai}}, \bibinfo {author} {\bibfnamefont {L.~V.}\ \bibnamefont
  {Thanayankizil}}, \bibinfo {author} {\bibfnamefont {D.~P.}\ \bibnamefont
  {Seetharam}},\ and\ \bibinfo {author} {\bibfnamefont {D.}~\bibnamefont
  {Chakraborty}},\ }\bibfield  {title} {\bibinfo {title} {Occupancy detection
  in commercial buildings using opportunistic context sources},\ }in\
  \href@noop {} {\emph {\bibinfo {booktitle} {2012 IEEE international
  conference on pervasive computing and communications workshops}}}\ (\bibinfo
  {organization} {IEEE},\ \bibinfo {year} {2012})\ pp.\ \bibinfo {pages}
  {463--466}\BibitemShut {NoStop}%
\bibitem [{\citenamefont {Fiebig}\ \emph {et~al.}(2017)\citenamefont {Fiebig},
  \citenamefont {Kochanneck}, \citenamefont {Mauser},\ and\ \citenamefont
  {Schmeck}}]{fiebig2017detecting}%
  \BibitemOpen
  \bibfield  {author} {\bibinfo {author} {\bibfnamefont {F.}~\bibnamefont
  {Fiebig}}, \bibinfo {author} {\bibfnamefont {S.}~\bibnamefont {Kochanneck}},
  \bibinfo {author} {\bibfnamefont {I.}~\bibnamefont {Mauser}},\ and\ \bibinfo
  {author} {\bibfnamefont {H.}~\bibnamefont {Schmeck}},\ }\bibfield  {title}
  {\bibinfo {title} {Detecting occupancy in smart buildings by data fusion from
  low-cost sensors: poster description},\ }in\ \href@noop {} {\emph {\bibinfo
  {booktitle} {Proceedings of the Eighth International Conference on Future
  Energy Systems}}}\ (\bibinfo {year} {2017})\ pp.\ \bibinfo {pages}
  {259--261}\BibitemShut {NoStop}%
\bibitem [{\citenamefont {Szczurek}\ \emph {et~al.}(2017)\citenamefont
  {Szczurek}, \citenamefont {Maciejewska},\ and\ \citenamefont
  {Pietrucha}}]{szczurek2017occupancy}%
  \BibitemOpen
  \bibfield  {author} {\bibinfo {author} {\bibfnamefont {A.}~\bibnamefont
  {Szczurek}}, \bibinfo {author} {\bibfnamefont {M.}~\bibnamefont
  {Maciejewska}},\ and\ \bibinfo {author} {\bibfnamefont {T.}~\bibnamefont
  {Pietrucha}},\ }\bibfield  {title} {\bibinfo {title} {Occupancy determination
  based on time series of co2 concentration, temperature and relative
  humidity},\ }\href@noop {} {\bibfield  {journal} {\bibinfo  {journal} {Energy
  and Buildings}\ }\textbf {\bibinfo {volume} {147}},\ \bibinfo {pages} {142}
  (\bibinfo {year} {2017})}\BibitemShut {NoStop}%
\bibitem [{\citenamefont {Raykov}\ \emph {et~al.}(2016)\citenamefont {Raykov},
  \citenamefont {Ozer}, \citenamefont {Dasika}, \citenamefont {Boukouvalas},\
  and\ \citenamefont {Little}}]{raykov2016predicting}%
  \BibitemOpen
  \bibfield  {author} {\bibinfo {author} {\bibfnamefont {Y.~P.}\ \bibnamefont
  {Raykov}}, \bibinfo {author} {\bibfnamefont {E.}~\bibnamefont {Ozer}},
  \bibinfo {author} {\bibfnamefont {G.}~\bibnamefont {Dasika}}, \bibinfo
  {author} {\bibfnamefont {A.}~\bibnamefont {Boukouvalas}},\ and\ \bibinfo
  {author} {\bibfnamefont {M.~A.}\ \bibnamefont {Little}},\ }\bibfield  {title}
  {\bibinfo {title} {Predicting room occupancy with a single passive infrared
  (pir) sensor through behavior extraction},\ }in\ \href@noop {} {\emph
  {\bibinfo {booktitle} {Proceedings of the 2016 ACM International Joint
  Conference on Pervasive and Ubiquitous Computing}}}\ (\bibinfo {year}
  {2016})\ pp.\ \bibinfo {pages} {1016--1027}\BibitemShut {NoStop}%
\bibitem [{\citenamefont {Yang}\ \emph {et~al.}(2014)\citenamefont {Yang},
  \citenamefont {Li}, \citenamefont {Becerik-Gerber},\ and\ \citenamefont
  {Orosz}}]{yang2014systematic}%
  \BibitemOpen
  \bibfield  {author} {\bibinfo {author} {\bibfnamefont {Z.}~\bibnamefont
  {Yang}}, \bibinfo {author} {\bibfnamefont {N.}~\bibnamefont {Li}}, \bibinfo
  {author} {\bibfnamefont {B.}~\bibnamefont {Becerik-Gerber}},\ and\ \bibinfo
  {author} {\bibfnamefont {M.}~\bibnamefont {Orosz}},\ }\bibfield  {title}
  {\bibinfo {title} {A systematic approach to occupancy modeling in ambient
  sensor-rich buildings},\ }\href@noop {} {\bibfield  {journal} {\bibinfo
  {journal} {Simulation}\ }\textbf {\bibinfo {volume} {90}},\ \bibinfo {pages}
  {960} (\bibinfo {year} {2014})}\BibitemShut {NoStop}%
\bibitem [{\citenamefont {Candanedo}\ and\ \citenamefont
  {Feldheim}(2016)}]{candanedo2016accurate}%
  \BibitemOpen
  \bibfield  {author} {\bibinfo {author} {\bibfnamefont {L.~M.}\ \bibnamefont
  {Candanedo}}\ and\ \bibinfo {author} {\bibfnamefont {V.}~\bibnamefont
  {Feldheim}},\ }\bibfield  {title} {\bibinfo {title} {Accurate occupancy
  detection of an office room from light, temperature, humidity and co2
  measurements using statistical learning models},\ }\href@noop {} {\bibfield
  {journal} {\bibinfo  {journal} {Energy and Buildings}\ }\textbf {\bibinfo
  {volume} {112}},\ \bibinfo {pages} {28} (\bibinfo {year} {2016})}\BibitemShut
  {NoStop}%
\bibitem [{\citenamefont {Arief-Ang}\ \emph
  {et~al.}(2017{\natexlab{b}})\citenamefont {Arief-Ang}, \citenamefont
  {Salim},\ and\ \citenamefont {Hamilton}}]{arief2017hoc}%
  \BibitemOpen
  \bibfield  {author} {\bibinfo {author} {\bibfnamefont {I.~B.}\ \bibnamefont
  {Arief-Ang}}, \bibinfo {author} {\bibfnamefont {F.~D.}\ \bibnamefont
  {Salim}},\ and\ \bibinfo {author} {\bibfnamefont {M.}~\bibnamefont
  {Hamilton}},\ }\bibfield  {title} {\bibinfo {title} {Da-hoc: semi-supervised
  domain adaptation for room occupancy prediction using co2 sensor data},\ }in\
  \href@noop {} {\emph {\bibinfo {booktitle} {Proceedings of the 4th ACM
  International Conference on Systems for Energy-Efficient Built
  Environments}}}\ (\bibinfo {year} {2017})\ pp.\ \bibinfo {pages}
  {1--10}\BibitemShut {NoStop}%
\bibitem [{\citenamefont {Peng}\ \emph {et~al.}(2018)\citenamefont {Peng},
  \citenamefont {Rysanek}, \citenamefont {Nagy},\ and\ \citenamefont
  {Schl{\"u}ter}}]{peng2018using}%
  \BibitemOpen
  \bibfield  {author} {\bibinfo {author} {\bibfnamefont {Y.}~\bibnamefont
  {Peng}}, \bibinfo {author} {\bibfnamefont {A.}~\bibnamefont {Rysanek}},
  \bibinfo {author} {\bibfnamefont {Z.}~\bibnamefont {Nagy}},\ and\ \bibinfo
  {author} {\bibfnamefont {A.}~\bibnamefont {Schl{\"u}ter}},\ }\bibfield
  {title} {\bibinfo {title} {Using machine learning techniques for
  occupancy-prediction-based cooling control in office buildings},\ }\href@noop
  {} {\bibfield  {journal} {\bibinfo  {journal} {Applied energy}\ }\textbf
  {\bibinfo {volume} {211}},\ \bibinfo {pages} {1343} (\bibinfo {year}
  {2018})}\BibitemShut {NoStop}%
\bibitem [{\citenamefont {Song}\ \emph {et~al.}(2019)\citenamefont {Song},
  \citenamefont {Niu}, \citenamefont {Lyu}, \citenamefont {Lyu},\ and\
  \citenamefont {Tian}}]{song2019time}%
  \BibitemOpen
  \bibfield  {author} {\bibinfo {author} {\bibfnamefont {L.}~\bibnamefont
  {Song}}, \bibinfo {author} {\bibfnamefont {X.}~\bibnamefont {Niu}}, \bibinfo
  {author} {\bibfnamefont {Q.}~\bibnamefont {Lyu}}, \bibinfo {author}
  {\bibfnamefont {S.}~\bibnamefont {Lyu}},\ and\ \bibinfo {author}
  {\bibfnamefont {T.}~\bibnamefont {Tian}},\ }\bibfield  {title} {\bibinfo
  {title} {A time-aware method for occupancy detection in a building},\ }in\
  \href@noop {} {\emph {\bibinfo {booktitle} {12th EAI International Conference
  on Mobile Multimedia Communications, Mobimedia 2019}}}\ (\bibinfo
  {organization} {European Alliance for Innovation (EAI)},\ \bibinfo {year}
  {2019})\BibitemShut {NoStop}%
\bibitem [{\citenamefont {Wolf}\ \emph {et~al.}(2019)\citenamefont {Wolf},
  \citenamefont {M{\o}ller}, \citenamefont {Bitsch}, \citenamefont {Krogstie},\
  and\ \citenamefont {Madsen}}]{sebastinwolf2019markov}%
  \BibitemOpen
  \bibfield  {author} {\bibinfo {author} {\bibfnamefont {S.}~\bibnamefont
  {Wolf}}, \bibinfo {author} {\bibfnamefont {J.~K.}\ \bibnamefont {M{\o}ller}},
  \bibinfo {author} {\bibfnamefont {M.~A.}\ \bibnamefont {Bitsch}}, \bibinfo
  {author} {\bibfnamefont {J.}~\bibnamefont {Krogstie}},\ and\ \bibinfo
  {author} {\bibfnamefont {H.}~\bibnamefont {Madsen}},\ }\bibfield  {title}
  {\bibinfo {title} {A markov-switching model for building occupant activity
  estimation},\ }\href@noop {} {\bibfield  {journal} {\bibinfo  {journal}
  {Energy and Buildings}\ }\textbf {\bibinfo {volume} {183}},\ \bibinfo {pages}
  {672} (\bibinfo {year} {2019})}\BibitemShut {NoStop}%
\bibitem [{\citenamefont {Depatla}\ \emph {et~al.}(2015)\citenamefont
  {Depatla}, \citenamefont {Muralidharan},\ and\ \citenamefont
  {Mostofi}}]{depatla2015occupancy}%
  \BibitemOpen
  \bibfield  {author} {\bibinfo {author} {\bibfnamefont {S.}~\bibnamefont
  {Depatla}}, \bibinfo {author} {\bibfnamefont {A.}~\bibnamefont
  {Muralidharan}},\ and\ \bibinfo {author} {\bibfnamefont {Y.}~\bibnamefont
  {Mostofi}},\ }\bibfield  {title} {\bibinfo {title} {Occupancy estimation
  using only wifi power measurements},\ }\href@noop {} {\bibfield  {journal}
  {\bibinfo  {journal} {IEEE Journal on Selected Areas in Communications}\
  }\textbf {\bibinfo {volume} {33}},\ \bibinfo {pages} {1381} (\bibinfo {year}
  {2015})}\BibitemShut {NoStop}%
\bibitem [{\citenamefont {Jain}\ \emph {et~al.}(2016)\citenamefont {Jain},
  \citenamefont {Chandan}, \citenamefont {Kumar}, \citenamefont {Arya},
  \citenamefont {Sridhar},\ and\ \citenamefont {Ramesh}}]{jain2016software}%
  \BibitemOpen
  \bibfield  {author} {\bibinfo {author} {\bibfnamefont {M.}~\bibnamefont
  {Jain}}, \bibinfo {author} {\bibfnamefont {V.}~\bibnamefont {Chandan}},
  \bibinfo {author} {\bibfnamefont {A.~P.}\ \bibnamefont {Kumar}}, \bibinfo
  {author} {\bibfnamefont {V.}~\bibnamefont {Arya}}, \bibinfo {author}
  {\bibfnamefont {R.}~\bibnamefont {Sridhar}},\ and\ \bibinfo {author}
  {\bibfnamefont {B.}~\bibnamefont {Ramesh}},\ }\bibfield  {title} {\bibinfo
  {title} {Software-only occupancy inference in a workplace findings from a
  field trial},\ }in\ \href@noop {} {\emph {\bibinfo {booktitle} {2016 IEEE
  Power \& Energy Society Innovative Smart Grid Technologies Conference
  (ISGT)}}}\ (\bibinfo {organization} {IEEE},\ \bibinfo {year} {2016})\ pp.\
  \bibinfo {pages} {1--5}\BibitemShut {NoStop}%
\bibitem [{\citenamefont {Candanedo}\ \emph
  {et~al.}(2017{\natexlab{a}})\citenamefont {Candanedo}, \citenamefont
  {Feldheim},\ and\ \citenamefont {Deramaix}}]{candanedo2017methodology}%
  \BibitemOpen
  \bibfield  {author} {\bibinfo {author} {\bibfnamefont {L.~M.}\ \bibnamefont
  {Candanedo}}, \bibinfo {author} {\bibfnamefont {V.}~\bibnamefont
  {Feldheim}},\ and\ \bibinfo {author} {\bibfnamefont {D.}~\bibnamefont
  {Deramaix}},\ }\bibfield  {title} {\bibinfo {title} {A methodology based on
  hidden markov models for occupancy detection and a case study in a low energy
  residential building},\ }\href@noop {} {\bibfield  {journal} {\bibinfo
  {journal} {Energy and Buildings}\ }\textbf {\bibinfo {volume} {148}},\
  \bibinfo {pages} {327} (\bibinfo {year} {2017}{\natexlab{a}})}\BibitemShut
  {NoStop}%
\bibitem [{\citenamefont {Pedersen}\ \emph {et~al.}(2017)\citenamefont
  {Pedersen}, \citenamefont {Nielsen},\ and\ \citenamefont
  {Petersen}}]{pedersen2017method}%
  \BibitemOpen
  \bibfield  {author} {\bibinfo {author} {\bibfnamefont {T.~H.}\ \bibnamefont
  {Pedersen}}, \bibinfo {author} {\bibfnamefont {K.~U.}\ \bibnamefont
  {Nielsen}},\ and\ \bibinfo {author} {\bibfnamefont {S.}~\bibnamefont
  {Petersen}},\ }\bibfield  {title} {\bibinfo {title} {Method for room
  occupancy detection based on trajectory of indoor climate sensor data},\
  }\href@noop {} {\bibfield  {journal} {\bibinfo  {journal} {Building and
  Environment}\ }\textbf {\bibinfo {volume} {115}},\ \bibinfo {pages} {147}
  (\bibinfo {year} {2017})}\BibitemShut {NoStop}%
\bibitem [{\citenamefont {Chen}\ \emph
  {et~al.}(2017{\natexlab{a}})\citenamefont {Chen}, \citenamefont {Zhu},
  \citenamefont {Masood},\ and\ \citenamefont {Soh}}]{chen2017environmental}%
  \BibitemOpen
  \bibfield  {author} {\bibinfo {author} {\bibfnamefont {Z.}~\bibnamefont
  {Chen}}, \bibinfo {author} {\bibfnamefont {Q.}~\bibnamefont {Zhu}}, \bibinfo
  {author} {\bibfnamefont {M.~K.}\ \bibnamefont {Masood}},\ and\ \bibinfo
  {author} {\bibfnamefont {Y.~C.}\ \bibnamefont {Soh}},\ }\bibfield  {title}
  {\bibinfo {title} {Environmental sensors-based occupancy estimation in
  buildings via ihmm-mlr},\ }\href@noop {} {\bibfield  {journal} {\bibinfo
  {journal} {IEEE Transactions on Industrial Informatics}\ }\textbf {\bibinfo
  {volume} {13}},\ \bibinfo {pages} {2184} (\bibinfo {year}
  {2017}{\natexlab{a}})}\BibitemShut {NoStop}%
\bibitem [{\citenamefont {Ryu}\ and\ \citenamefont
  {Moon}(2016)}]{ryu2016development}%
  \BibitemOpen
  \bibfield  {author} {\bibinfo {author} {\bibfnamefont {S.~H.}\ \bibnamefont
  {Ryu}}\ and\ \bibinfo {author} {\bibfnamefont {H.~J.}\ \bibnamefont {Moon}},\
  }\bibfield  {title} {\bibinfo {title} {Development of an occupancy prediction
  model using indoor environmental data based on machine learning techniques},\
  }\href@noop {} {\bibfield  {journal} {\bibinfo  {journal} {Building and
  Environment}\ }\textbf {\bibinfo {volume} {107}},\ \bibinfo {pages} {1}
  (\bibinfo {year} {2016})}\BibitemShut {NoStop}%
\bibitem [{\citenamefont {Chaney}\ \emph {et~al.}(2016)\citenamefont {Chaney},
  \citenamefont {Owens},\ and\ \citenamefont {Peacock}}]{chaney2016evidence}%
  \BibitemOpen
  \bibfield  {author} {\bibinfo {author} {\bibfnamefont {J.}~\bibnamefont
  {Chaney}}, \bibinfo {author} {\bibfnamefont {E.~H.}\ \bibnamefont {Owens}},\
  and\ \bibinfo {author} {\bibfnamefont {A.~D.}\ \bibnamefont {Peacock}},\
  }\bibfield  {title} {\bibinfo {title} {An evidence based approach to
  determining residential occupancy and its role in demand response
  management},\ }\href@noop {} {\bibfield  {journal} {\bibinfo  {journal}
  {Energy and Buildings}\ }\textbf {\bibinfo {volume} {125}},\ \bibinfo {pages}
  {254} (\bibinfo {year} {2016})}\BibitemShut {NoStop}%
\bibitem [{\citenamefont {Chen}\ \emph
  {et~al.}(2017{\natexlab{b}})\citenamefont {Chen}, \citenamefont {Zhao},
  \citenamefont {Zhu}, \citenamefont {Masood}, \citenamefont {Soh},\ and\
  \citenamefont {Mao}}]{chen2017building}%
  \BibitemOpen
  \bibfield  {author} {\bibinfo {author} {\bibfnamefont {Z.}~\bibnamefont
  {Chen}}, \bibinfo {author} {\bibfnamefont {R.}~\bibnamefont {Zhao}}, \bibinfo
  {author} {\bibfnamefont {Q.}~\bibnamefont {Zhu}}, \bibinfo {author}
  {\bibfnamefont {M.~K.}\ \bibnamefont {Masood}}, \bibinfo {author}
  {\bibfnamefont {Y.~C.}\ \bibnamefont {Soh}},\ and\ \bibinfo {author}
  {\bibfnamefont {K.}~\bibnamefont {Mao}},\ }\bibfield  {title} {\bibinfo
  {title} {Building occupancy estimation with environmental sensors via
  cdblstm},\ }\href@noop {} {\bibfield  {journal} {\bibinfo  {journal} {IEEE
  Transactions on Industrial Electronics}\ }\textbf {\bibinfo {volume} {64}},\
  \bibinfo {pages} {9549} (\bibinfo {year} {2017}{\natexlab{b}})}\BibitemShut
  {NoStop}%
\bibitem [{\citenamefont {Pan}\ \emph {et~al.}(2014)\citenamefont {Pan},
  \citenamefont {Bonde}, \citenamefont {Jing}, \citenamefont {Zhang},
  \citenamefont {Zhang},\ and\ \citenamefont {Noh}}]{pan2014boes}%
  \BibitemOpen
  \bibfield  {author} {\bibinfo {author} {\bibfnamefont {S.}~\bibnamefont
  {Pan}}, \bibinfo {author} {\bibfnamefont {A.}~\bibnamefont {Bonde}}, \bibinfo
  {author} {\bibfnamefont {J.}~\bibnamefont {Jing}}, \bibinfo {author}
  {\bibfnamefont {L.}~\bibnamefont {Zhang}}, \bibinfo {author} {\bibfnamefont
  {P.}~\bibnamefont {Zhang}},\ and\ \bibinfo {author} {\bibfnamefont {H.~Y.}\
  \bibnamefont {Noh}},\ }\bibfield  {title} {\bibinfo {title} {Boes: building
  occupancy estimation system using sparse ambient vibration monitoring},\ }in\
  \href@noop {} {\emph {\bibinfo {booktitle} {Sensors and Smart Structures
  Technologies for Civil, Mechanical, and Aerospace Systems 2014}}},\ Vol.\
  \bibinfo {volume} {9061}\ (\bibinfo {organization} {International Society for
  Optics and Photonics},\ \bibinfo {year} {2014})\ p.\ \bibinfo {pages}
  {90611O}\BibitemShut {NoStop}%
\bibitem [{\citenamefont {Barbour}\ \emph {et~al.}(2019)\citenamefont
  {Barbour}, \citenamefont {Davila}, \citenamefont {Gupta}, \citenamefont
  {Reinhart}, \citenamefont {Kaur},\ and\ \citenamefont
  {Gonz{\'a}lez}}]{barbour2019planning}%
  \BibitemOpen
  \bibfield  {author} {\bibinfo {author} {\bibfnamefont {E.}~\bibnamefont
  {Barbour}}, \bibinfo {author} {\bibfnamefont {C.~C.}\ \bibnamefont {Davila}},
  \bibinfo {author} {\bibfnamefont {S.}~\bibnamefont {Gupta}}, \bibinfo
  {author} {\bibfnamefont {C.}~\bibnamefont {Reinhart}}, \bibinfo {author}
  {\bibfnamefont {J.}~\bibnamefont {Kaur}},\ and\ \bibinfo {author}
  {\bibfnamefont {M.~C.}\ \bibnamefont {Gonz{\'a}lez}},\ }\bibfield  {title}
  {\bibinfo {title} {Planning for sustainable cities by estimating building
  occupancy with mobile phones},\ }\href@noop {} {\bibfield  {journal}
  {\bibinfo  {journal} {Nature communications}\ }\textbf {\bibinfo {volume}
  {10}},\ \bibinfo {pages} {1} (\bibinfo {year} {2019})}\BibitemShut {NoStop}%
\bibitem [{\citenamefont {Jacob}\ \emph {et~al.}(2010)\citenamefont {Jacob},
  \citenamefont {Burhenne}, \citenamefont {Florita},\ and\ \citenamefont
  {Henze}}]{jacob2010optimizing}%
  \BibitemOpen
  \bibfield  {author} {\bibinfo {author} {\bibfnamefont {D.}~\bibnamefont
  {Jacob}}, \bibinfo {author} {\bibfnamefont {S.}~\bibnamefont {Burhenne}},
  \bibinfo {author} {\bibfnamefont {A.~R.}\ \bibnamefont {Florita}},\ and\
  \bibinfo {author} {\bibfnamefont {G.~P.}\ \bibnamefont {Henze}},\ }\bibfield
  {title} {\bibinfo {title} {Optimizing building energy simulation models in
  the face of uncertainty},\ }\href@noop {} {\bibfield  {journal} {\bibinfo
  {journal} {Proceedings of SimBuild}\ }\textbf {\bibinfo {volume} {4}},\
  \bibinfo {pages} {118} (\bibinfo {year} {2010})}\BibitemShut {NoStop}%
\bibitem [{\citenamefont {Hu}\ \emph {et~al.}(2016)\citenamefont {Hu},
  \citenamefont {Oldewurtel}, \citenamefont {Balandat}, \citenamefont
  {Vrettos}, \citenamefont {Zhou},\ and\ \citenamefont
  {Tomlin}}]{hu2016building}%
  \BibitemOpen
  \bibfield  {author} {\bibinfo {author} {\bibfnamefont {Q.}~\bibnamefont
  {Hu}}, \bibinfo {author} {\bibfnamefont {F.}~\bibnamefont {Oldewurtel}},
  \bibinfo {author} {\bibfnamefont {M.}~\bibnamefont {Balandat}}, \bibinfo
  {author} {\bibfnamefont {E.}~\bibnamefont {Vrettos}}, \bibinfo {author}
  {\bibfnamefont {D.}~\bibnamefont {Zhou}},\ and\ \bibinfo {author}
  {\bibfnamefont {C.~J.}\ \bibnamefont {Tomlin}},\ }\bibfield  {title}
  {\bibinfo {title} {Building model identification during regular
  operation-empirical results and challenges},\ }in\ \href@noop {} {\emph
  {\bibinfo {booktitle} {2016 American Control Conference (ACC)}}}\ (\bibinfo
  {organization} {IEEE},\ \bibinfo {year} {2016})\ pp.\ \bibinfo {pages}
  {605--610}\BibitemShut {NoStop}%
\bibitem [{\citenamefont {Burger}\ and\ \citenamefont
  {Moura}(2016)}]{burger2016recursive}%
  \BibitemOpen
  \bibfield  {author} {\bibinfo {author} {\bibfnamefont {E.~M.}\ \bibnamefont
  {Burger}}\ and\ \bibinfo {author} {\bibfnamefont {S.~J.}\ \bibnamefont
  {Moura}},\ }\bibfield  {title} {\bibinfo {title} {Recursive parameter
  estimation of thermostatically controlled loads via unscented kalman
  filter},\ }\href@noop {} {\bibfield  {journal} {\bibinfo  {journal}
  {Sustainable Energy, Grids and Networks}\ }\textbf {\bibinfo {volume} {8}},\
  \bibinfo {pages} {12} (\bibinfo {year} {2016})}\BibitemShut {NoStop}%
\bibitem [{\citenamefont {Ji}\ \emph {et~al.}(2016)\citenamefont {Ji},
  \citenamefont {Xu}, \citenamefont {Duan},\ and\ \citenamefont
  {Lu}}]{ji2016estimating}%
  \BibitemOpen
  \bibfield  {author} {\bibinfo {author} {\bibfnamefont {Y.}~\bibnamefont
  {Ji}}, \bibinfo {author} {\bibfnamefont {P.}~\bibnamefont {Xu}}, \bibinfo
  {author} {\bibfnamefont {P.}~\bibnamefont {Duan}},\ and\ \bibinfo {author}
  {\bibfnamefont {X.}~\bibnamefont {Lu}},\ }\bibfield  {title} {\bibinfo
  {title} {Estimating hourly cooling load in commercial buildings using a
  thermal network model and electricity submetering data},\ }\href@noop {}
  {\bibfield  {journal} {\bibinfo  {journal} {Applied Energy}\ }\textbf
  {\bibinfo {volume} {169}},\ \bibinfo {pages} {309} (\bibinfo {year}
  {2016})}\BibitemShut {NoStop}%
\bibitem [{\citenamefont {Dewson}\ \emph {et~al.}(1993)\citenamefont {Dewson},
  \citenamefont {Day},\ and\ \citenamefont {Irving}}]{dewson1993least}%
  \BibitemOpen
  \bibfield  {author} {\bibinfo {author} {\bibfnamefont {T.}~\bibnamefont
  {Dewson}}, \bibinfo {author} {\bibfnamefont {B.}~\bibnamefont {Day}},\ and\
  \bibinfo {author} {\bibfnamefont {A.}~\bibnamefont {Irving}},\ }\bibfield
  {title} {\bibinfo {title} {Least squares parameter estimation of a reduced
  order thermal model of an experimental building},\ }\href@noop {} {\bibfield
  {journal} {\bibinfo  {journal} {Building and Environment}\ }\textbf {\bibinfo
  {volume} {28}},\ \bibinfo {pages} {127} (\bibinfo {year} {1993})}\BibitemShut
  {NoStop}%
\bibitem [{\citenamefont {Sangogboye}\ \emph {et~al.}(2017)\citenamefont
  {Sangogboye}, \citenamefont {Arendt}, \citenamefont {Singh}, \citenamefont
  {Veje}, \citenamefont {Kj{\ae}rgaard},\ and\ \citenamefont
  {J{\o}rgensen}}]{sangogboye2017performance}%
  \BibitemOpen
  \bibfield  {author} {\bibinfo {author} {\bibfnamefont {F.~C.}\ \bibnamefont
  {Sangogboye}}, \bibinfo {author} {\bibfnamefont {K.}~\bibnamefont {Arendt}},
  \bibinfo {author} {\bibfnamefont {A.}~\bibnamefont {Singh}}, \bibinfo
  {author} {\bibfnamefont {C.~T.}\ \bibnamefont {Veje}}, \bibinfo {author}
  {\bibfnamefont {M.~B.}\ \bibnamefont {Kj{\ae}rgaard}},\ and\ \bibinfo
  {author} {\bibfnamefont {B.~N.}\ \bibnamefont {J{\o}rgensen}},\ }\bibfield
  {title} {\bibinfo {title} {Performance comparison of occupancy count
  estimation and prediction with common versus dedicated sensors for building
  model predictive control},\ }in\ \href@noop {} {\emph {\bibinfo {booktitle}
  {Building Simulation}}},\ Vol.~\bibinfo {volume} {10}\ (\bibinfo
  {organization} {Springer},\ \bibinfo {year} {2017})\ pp.\ \bibinfo {pages}
  {829--843}\BibitemShut {NoStop}%
\bibitem [{\citenamefont {Gruber}\ \emph {et~al.}(2014)\citenamefont {Gruber},
  \citenamefont {Tr{\"u}schel},\ and\ \citenamefont
  {Dalenb{\"a}ck}}]{gruber2014co2}%
  \BibitemOpen
  \bibfield  {author} {\bibinfo {author} {\bibfnamefont {M.}~\bibnamefont
  {Gruber}}, \bibinfo {author} {\bibfnamefont {A.}~\bibnamefont
  {Tr{\"u}schel}},\ and\ \bibinfo {author} {\bibfnamefont {J.-O.}\ \bibnamefont
  {Dalenb{\"a}ck}},\ }\bibfield  {title} {\bibinfo {title} {Co2 sensors for
  occupancy estimations: Potential in building automation applications},\
  }\href@noop {} {\bibfield  {journal} {\bibinfo  {journal} {Energy and
  Buildings}\ }\textbf {\bibinfo {volume} {84}},\ \bibinfo {pages} {548}
  (\bibinfo {year} {2014})}\BibitemShut {NoStop}%
\bibitem [{\citenamefont {Ebadat}\ \emph {et~al.}(2013)\citenamefont {Ebadat},
  \citenamefont {Bottegal}, \citenamefont {Varagnolo}, \citenamefont
  {Wahlberg},\ and\ \citenamefont {Johansson}}]{ebadat2013estimation}%
  \BibitemOpen
  \bibfield  {author} {\bibinfo {author} {\bibfnamefont {A.}~\bibnamefont
  {Ebadat}}, \bibinfo {author} {\bibfnamefont {G.}~\bibnamefont {Bottegal}},
  \bibinfo {author} {\bibfnamefont {D.}~\bibnamefont {Varagnolo}}, \bibinfo
  {author} {\bibfnamefont {B.}~\bibnamefont {Wahlberg}},\ and\ \bibinfo
  {author} {\bibfnamefont {K.~H.}\ \bibnamefont {Johansson}},\ }\bibfield
  {title} {\bibinfo {title} {Estimation of building occupancy levels through
  environmental signals deconvolution},\ }in\ \href@noop {} {\emph {\bibinfo
  {booktitle} {Proceedings of the 5th ACM Workshop on Embedded Systems For
  Energy-Efficient Buildings}}}\ (\bibinfo {year} {2013})\ pp.\ \bibinfo
  {pages} {1--8}\BibitemShut {NoStop}%
\bibitem [{\citenamefont {Jung}\ and\ \citenamefont
  {Jazizadeh}(2019)}]{jung2019human}%
  \BibitemOpen
  \bibfield  {author} {\bibinfo {author} {\bibfnamefont {W.}~\bibnamefont
  {Jung}}\ and\ \bibinfo {author} {\bibfnamefont {F.}~\bibnamefont
  {Jazizadeh}},\ }\bibfield  {title} {\bibinfo {title} {Human-in-the-loop hvac
  operations: A quantitative review on occupancy, comfort, and
  energy-efficiency dimensions},\ }\href@noop {} {\bibfield  {journal}
  {\bibinfo  {journal} {Applied Energy}\ }\textbf {\bibinfo {volume} {239}},\
  \bibinfo {pages} {1471} (\bibinfo {year} {2019})}\BibitemShut {NoStop}%
\bibitem [{\citenamefont {Chen}\ \emph {et~al.}(2018)\citenamefont {Chen},
  \citenamefont {Jiang},\ and\ \citenamefont {Xie}}]{chen2018building}%
  \BibitemOpen
  \bibfield  {author} {\bibinfo {author} {\bibfnamefont {Z.}~\bibnamefont
  {Chen}}, \bibinfo {author} {\bibfnamefont {C.}~\bibnamefont {Jiang}},\ and\
  \bibinfo {author} {\bibfnamefont {L.}~\bibnamefont {Xie}},\ }\bibfield
  {title} {\bibinfo {title} {Building occupancy estimation and detection: A
  review},\ }\href@noop {} {\bibfield  {journal} {\bibinfo  {journal} {Energy
  and Buildings}\ }\textbf {\bibinfo {volume} {169}},\ \bibinfo {pages} {260}
  (\bibinfo {year} {2018})}\BibitemShut {NoStop}%
\bibitem [{\citenamefont {Lokhov}\ \emph {et~al.}(2018)\citenamefont {Lokhov},
  \citenamefont {Vuffray}, \citenamefont {Shemetov}, \citenamefont {Deka},\
  and\ \citenamefont {Chertkov}}]{lokhov2018online}%
  \BibitemOpen
  \bibfield  {author} {\bibinfo {author} {\bibfnamefont {A.~Y.}\ \bibnamefont
  {Lokhov}}, \bibinfo {author} {\bibfnamefont {M.}~\bibnamefont {Vuffray}},
  \bibinfo {author} {\bibfnamefont {D.}~\bibnamefont {Shemetov}}, \bibinfo
  {author} {\bibfnamefont {D.}~\bibnamefont {Deka}},\ and\ \bibinfo {author}
  {\bibfnamefont {M.}~\bibnamefont {Chertkov}},\ }\bibfield  {title} {\bibinfo
  {title} {Online learning of power transmission dynamics},\ }in\ \href@noop {}
  {\emph {\bibinfo {booktitle} {2018 Power Systems Computation Conference
  (PSCC)}}}\ (\bibinfo {organization} {IEEE},\ \bibinfo {year} {2018})\ pp.\
  \bibinfo {pages} {1--7}\BibitemShut {NoStop}%
\bibitem [{\citenamefont {Simchowitz}\ \emph {et~al.}(2018)\citenamefont
  {Simchowitz}, \citenamefont {Mania}, \citenamefont {Tu}, \citenamefont
  {Jordan},\ and\ \citenamefont {Recht}}]{simchowitz2018learning}%
  \BibitemOpen
  \bibfield  {author} {\bibinfo {author} {\bibfnamefont {M.}~\bibnamefont
  {Simchowitz}}, \bibinfo {author} {\bibfnamefont {H.}~\bibnamefont {Mania}},
  \bibinfo {author} {\bibfnamefont {S.}~\bibnamefont {Tu}}, \bibinfo {author}
  {\bibfnamefont {M.~I.}\ \bibnamefont {Jordan}},\ and\ \bibinfo {author}
  {\bibfnamefont {B.}~\bibnamefont {Recht}},\ }\bibfield  {title} {\bibinfo
  {title} {Learning without mixing: Towards a sharp analysis of linear system
  identification},\ }\href@noop {} {\bibfield  {journal} {\bibinfo  {journal}
  {arXiv preprint arXiv:1802.08334}\ } (\bibinfo {year} {2018})}\BibitemShut
  {NoStop}%
\bibitem [{\citenamefont {Sarkar}\ and\ \citenamefont
  {Rakhlin}(2019)}]{sarkar2019near}%
  \BibitemOpen
  \bibfield  {author} {\bibinfo {author} {\bibfnamefont {T.}~\bibnamefont
  {Sarkar}}\ and\ \bibinfo {author} {\bibfnamefont {A.}~\bibnamefont
  {Rakhlin}},\ }\bibfield  {title} {\bibinfo {title} {Near optimal finite time
  identification of arbitrary linear dynamical systems},\ }in\ \href@noop {}
  {\emph {\bibinfo {booktitle} {International Conference on Machine
  Learning}}}\ (\bibinfo {year} {2019})\ pp.\ \bibinfo {pages}
  {5610--5618}\BibitemShut {NoStop}%
\bibitem [{\citenamefont {Bell}\ \emph {et~al.}(2013)\citenamefont {Bell},
  \citenamefont {Palecki}, \citenamefont {Baker}, \citenamefont {Collins},
  \citenamefont {Lawrimore}, \citenamefont {Leeper}, \citenamefont {Hall},
  \citenamefont {Kochendorfer}, \citenamefont {Meyers}, \citenamefont {Wilson}
  \emph {et~al.}}]{bell2013us}%
  \BibitemOpen
  \bibfield  {author} {\bibinfo {author} {\bibfnamefont {J.~E.}\ \bibnamefont
  {Bell}}, \bibinfo {author} {\bibfnamefont {M.~A.}\ \bibnamefont {Palecki}},
  \bibinfo {author} {\bibfnamefont {C.~B.}\ \bibnamefont {Baker}}, \bibinfo
  {author} {\bibfnamefont {W.~G.}\ \bibnamefont {Collins}}, \bibinfo {author}
  {\bibfnamefont {J.~H.}\ \bibnamefont {Lawrimore}}, \bibinfo {author}
  {\bibfnamefont {R.~D.}\ \bibnamefont {Leeper}}, \bibinfo {author}
  {\bibfnamefont {M.~E.}\ \bibnamefont {Hall}}, \bibinfo {author}
  {\bibfnamefont {J.}~\bibnamefont {Kochendorfer}}, \bibinfo {author}
  {\bibfnamefont {T.~P.}\ \bibnamefont {Meyers}}, \bibinfo {author}
  {\bibfnamefont {T.}~\bibnamefont {Wilson}}, \emph {et~al.},\ }\bibfield
  {title} {\bibinfo {title} {Us climate reference network soil moisture and
  temperature observations},\ }\href@noop {} {\bibfield  {journal} {\bibinfo
  {journal} {Journal of Hydrometeorology}\ }\textbf {\bibinfo {volume} {14}},\
  \bibinfo {pages} {977} (\bibinfo {year} {2013})}\BibitemShut {NoStop}%
\bibitem [{\citenamefont {Privara}\ \emph {et~al.}(2013)\citenamefont
  {Privara}, \citenamefont {Cigler}, \citenamefont {V{\'a}{\v{n}}a},
  \citenamefont {Oldewurtel}, \citenamefont {Sagerschnig},\ and\ \citenamefont
  {{\v{Z}}{\'a}{\v{c}}ekov{\'a}}}]{privara2013building}%
  \BibitemOpen
  \bibfield  {author} {\bibinfo {author} {\bibfnamefont {S.}~\bibnamefont
  {Privara}}, \bibinfo {author} {\bibfnamefont {J.}~\bibnamefont {Cigler}},
  \bibinfo {author} {\bibfnamefont {Z.}~\bibnamefont {V{\'a}{\v{n}}a}},
  \bibinfo {author} {\bibfnamefont {F.}~\bibnamefont {Oldewurtel}}, \bibinfo
  {author} {\bibfnamefont {C.}~\bibnamefont {Sagerschnig}},\ and\ \bibinfo
  {author} {\bibfnamefont {E.}~\bibnamefont {{\v{Z}}{\'a}{\v{c}}ekov{\'a}}},\
  }\bibfield  {title} {\bibinfo {title} {Building modeling as a crucial part
  for building predictive control},\ }\href@noop {} {\bibfield  {journal}
  {\bibinfo  {journal} {Energy and Buildings}\ }\textbf {\bibinfo {volume}
  {56}},\ \bibinfo {pages} {8} (\bibinfo {year} {2013})}\BibitemShut {NoStop}%
\bibitem [{\citenamefont {Hazyuk}\ \emph {et~al.}(2012)\citenamefont {Hazyuk},
  \citenamefont {Ghiaus},\ and\ \citenamefont {Penhouet}}]{hazyuk2012optimal}%
  \BibitemOpen
  \bibfield  {author} {\bibinfo {author} {\bibfnamefont {I.}~\bibnamefont
  {Hazyuk}}, \bibinfo {author} {\bibfnamefont {C.}~\bibnamefont {Ghiaus}},\
  and\ \bibinfo {author} {\bibfnamefont {D.}~\bibnamefont {Penhouet}},\
  }\bibfield  {title} {\bibinfo {title} {Optimal temperature control of
  intermittently heated buildings using model predictive control: Part
  i--building modeling},\ }\href@noop {} {\bibfield  {journal} {\bibinfo
  {journal} {Building and Environment}\ }\textbf {\bibinfo {volume} {51}},\
  \bibinfo {pages} {379} (\bibinfo {year} {2012})}\BibitemShut {NoStop}%
\bibitem [{\citenamefont {Candanedo}\ \emph
  {et~al.}(2017{\natexlab{b}})\citenamefont {Candanedo}, \citenamefont
  {Feldheim},\ and\ \citenamefont {Deramaix}}]{candanedo2017data}%
  \BibitemOpen
  \bibfield  {author} {\bibinfo {author} {\bibfnamefont {L.~M.}\ \bibnamefont
  {Candanedo}}, \bibinfo {author} {\bibfnamefont {V.}~\bibnamefont
  {Feldheim}},\ and\ \bibinfo {author} {\bibfnamefont {D.}~\bibnamefont
  {Deramaix}},\ }\bibfield  {title} {\bibinfo {title} {Data driven prediction
  models of energy use of appliances in a low-energy house},\ }\href@noop {}
  {\bibfield  {journal} {\bibinfo  {journal} {Energy and buildings}\ }\textbf
  {\bibinfo {volume} {140}},\ \bibinfo {pages} {81} (\bibinfo {year}
  {2017}{\natexlab{b}})}\BibitemShut {NoStop}%
\end{thebibliography}%


\providecommand{\noopsort}[1]{}\providecommand{\singleletter}[1]{#1}%
%

\end{document}